\documentclass{article}

\usepackage[preprint]{neurips_2025}
\usepackage{multirow}
\usepackage{subcaption}
\usepackage{enumitem}
\usepackage{graphicx}
\usepackage{amsmath}

\usepackage[utf8]{inputenc} 
\usepackage[T1]{fontenc}    
\usepackage{hyperref}       
\usepackage{url}            
\usepackage{booktabs}       
\usepackage{amsfonts}       
\usepackage{nicefrac}       
\usepackage{microtype}      
\usepackage{xcolor}         

\title{HunyuanVideo-HOMA: Generic \underline{H}uman-\underline{O}bject Interaction in \underline{M}ultimodal Driven Human \underline{A}nimation}

\author{%
  Ziyao Huang$^{12}$$^{*}$, Zixiang Zhou$^2$$\thanks{Equal contribution}$, Juan Cao$^1$, Yifeng Ma$^2$, Yi Chen$^2$, Zejing Rao$^1$,\\
  \textbf{Zhiyong Xu$^2$}, \textbf{Hongmei Wang$^2$}, \textbf{Qin Lin$^2$}, \textbf{Yuan Zhou$^2$}, \textbf{Qinglin Lu$^{2}$\thanks{Corresponding author (qinglinlu@tencent.com)}}, \textbf{Fan Tang$^{1}$}\thanks{Corresponding author (tfan.108@gmail.com)} \\
  $^1$University of Chinese Academy of Sciences    $^2$Tencent Hunyuan 
}

\begin{document}

\maketitle

\begin{figure*}[htbp]
  \includegraphics[width=\textwidth]{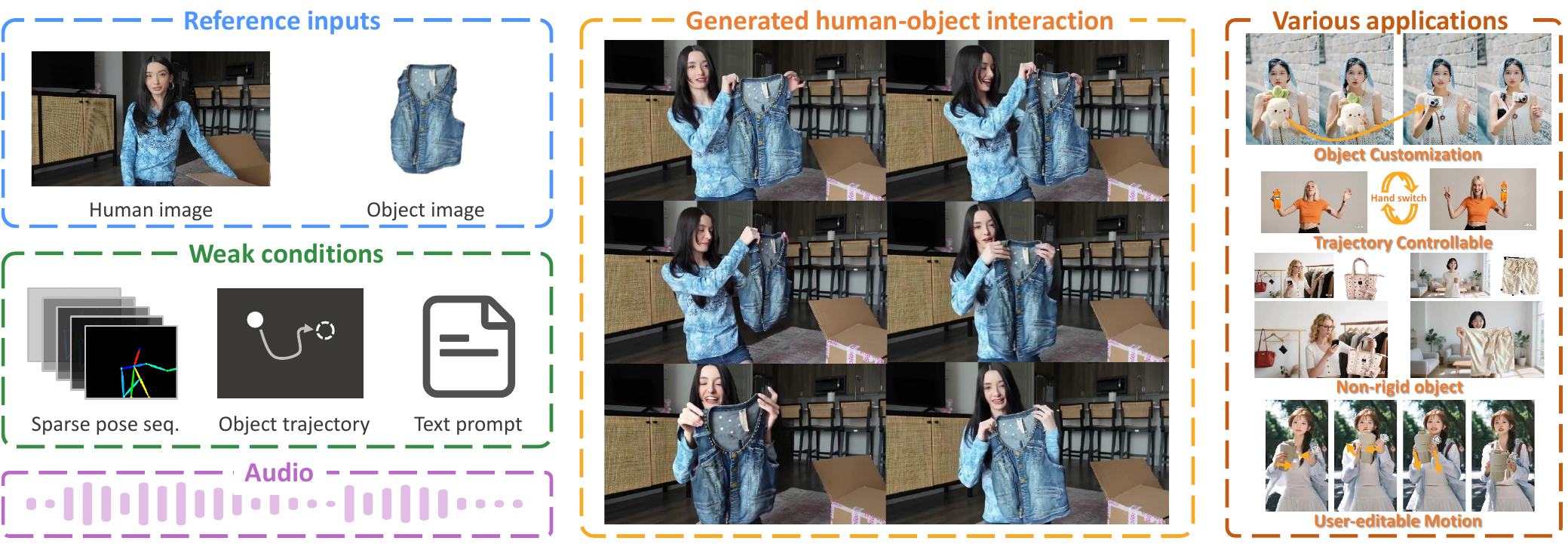}
  \caption{HunyuanVideo-HOMA is a multimodal-driven human animation framework for human-object interaction (HOI) generation, utilizing weak conditions to enable HOI video generation, delivering high-fidelity results from a single object input while maintaining strong generalizability across applications.}
  \label{fig:teaser}
\end{figure*}

\begin{abstract}
To address key limitations in human-object interaction (HOI) video generation—specifically the reliance on curated motion data, limited generalization to novel objects/scenarios, and restricted accessibility—we introduce HunyuanVideo-HOMA, a weakly conditioned multimodal-driven framework. HunyuanVideo-HOMA enhances controllability and reduces dependency on precise inputs through sparse, decoupled motion guidance. It encodes appearance and motion signals into the dual input space of a multimodal diffusion transformer (MMDiT), fusing them within a shared context space to synthesize temporally consistent and physically plausible interactions. To optimize training, we integrate a parameter-space HOI adapter initialized from pretrained MMDiT weights, preserving prior knowledge while enabling efficient adaptation, and a facial cross-attention adapter for anatomically accurate audio-driven lip synchronization. Extensive experiments confirm state-of-the-art performance in interaction naturalness and generalization under weak supervision. Finally, HunyuanVideo-HOMA demonstrates versatility in text-conditioned generation and interactive object manipulation, supported by a user-friendly demo interface. 
The project page is \url{https://bone-11.github.io/homa-page/}.
\end{abstract}

\section{Introduction}
\label{sec:introduction}
Human image animation for generating human-centric videos~\cite{hu2024animateanyone, wang2024disco, chang2023magicpose, xu2024magicanimate} has been extensively studied in recent years. 
Advancements in video diffusion models have markedly enhanced the quality of synthesized human motion, while concurrently enabling diverse driving modalities such as pose~\cite{hu2024animateanyone, zhang2024mimicmotion} and audio~\cite{zhang2023sadtalker, ji2024sonic, meng2024echomimicv2, lin2025omnihuman}. 
Building upon single-human-driven methodologies, recent breakthroughs have integrated human-object interaction (HOI) into these frameworks~\cite{xu2024anchorcrafter, xue2024hoi, pang2024manivideo, fan2025rehold}, enabling the generation of interactive objects alongside their corresponding motion dynamics. 
These approaches offer considerable potential for generating realistic human-object interactions in videos, a capability with broad applications across diverse domains, including online retail, digital advertising, and film production.

Despite the progress in HOI video generation, significant challenges persist in achieving robust generalization and applicability. A major limitation stems from the reliance on strong motion control mechanisms, such as actor-captured motion sequences~\cite{xu2024anchorcrafter, fan2025rehold, xue2024hoi} or computationally intensive 3D modeling~\cite{pang2024manivideo}. 
These requirements impose constraints on real-world usability. Furthermore, while existing methods demonstrate proficiency in handling simple rigid objects, their generalization to diverse object categories—including rigid objects with intricate geometries and surface properties, as well as deformable or articulated objects—remains understudied and inadequately addressed.

To address these challenges, we propose HunyuanVideo-HOMA, a novel framework built upon HunyuanVideo~\cite{kong2024hunyuanvideo}, designed to provide a more practical and flexible solution for HOI video generation through a multimodal-driven human animation pipeline. To mitigate the reliance on strongly supervised motion sequences, we explore a weak motion control paradigm that enhances both user editability and model adaptability. Specifically, we introduce a spatially sparse and decoupled motion condition, where only a single arm skeleton and the trajectory of the object’s center point are employed as minimal representations of human and object motion. This formulation captures essential interaction dynamics with the weakest possible control signal. Built upon this weak condition model, our framework enables the animation of a reference human and the seamless injection of interactive objects into the generated video, achieving controllable motion and natural interactions. Additionally, we incorporate text and audio inputs as complementary modalities, offering further control flexibility while maintaining the framework’s sparse input philosophy.

Under weak condition control, the model is expected to produce reasonable outputs despite missing substantial guidance. 
In such scenarios of weak guidance and strong inference, we argue that the model should be encouraged to leverage generative priors to fill in the uncontrolled components while still adhering to the given controls.
To address this issue, we introduce a novel approach built upon the input context space of multi-modal diffusion transformer (MMDiT)~\cite{esser2024sd3-mmdit} for controllable HOI generation. Specifically, we encode appearance and motion conditions into the context space of the MMDiT framework, ensuring precise and coherent integration of both visual and kinematic information. 
To enhance object appearance consistency and improve the plausibility of human-object interactions, we introduce an HOI Adapter, which is also in MMDiT architecture and initialized with the pretrained parameters from HunyuanVideo~\cite{kong2024hunyuanvideo}, leveraging prior knowledge to accelerate convergence and achieve superior consistency and interaction realism.
For lip-syncing control, we propose an Audio Cross-attention Adapter that injects speech signals into facial regions of generated videos, enabling temporally aligned and anatomically accurate lip movements.
By exploiting the model’s generative priors, our approach effectively enhances the consistency of object injection.
The data pipeline and training strategies are designed to optimize the performance of the HunyuanVideo-HOMA framework, including a depth-aware approach to filter out real-world video data without human-object interaction.

Experimental results demonstrate that HunyuanVideo-HOMA outperforms existing methods, with improved generalization capabilities of object appearance preservation and interaction naturalness. 
Various applications under different weak conditions are also demonstrated, including motion generation guided by different types of weak pose control, text-conditioned generation, and diverse modifications to object appearance and trajectory. Additionally, we present an interactive demo to support our weak-condition framework and facilitate user-guided refinement. In summary, our contributions are as follows:
\begin{itemize}[leftmargin=10pt]
  \item We propose HunyuanVideo-HOMA, a novel framework that introduces weak motion guidance to reduce the dependency on strong motion conditions for HOI video generation, achieving both flexibility and precision for real-world applications.
  \item Under limited weak conditions, we explore a novel solution to inject HOI appearance and motion in a manner that aligns with the model’s generative prior and enhances consistency in conditioning and naturalness in interaction generation.
  \item We conduct comprehensive quantitative and qualitative evaluations to validate the effectiveness of HunyuanVideo-HOMA, benchmarking it against state-of-the-art diffusion-based methods for HOI video generation and showcasing multiple application scenarios.
\end{itemize}

\section{Related Work}
\label{sec:related_work}

\subsection{Motion-controllable Human Video Generation}
With the advancement of video generation models~\cite{blattmann2023svd, kong2024hunyuanvideo, wang2025wan, zhou2024allegro}, the controllability of human subjects has become a central research focus. Key challenges include accurately controlling human motions using multimodal inputs such as pose and audio while preserving identity and appearance. 
For pose-driven control, the research trajectory includes methods from UNet-based approaches~\cite{hu2024animateanyone, zhu2024champ, wang2024humanvid, zhang2024mimicmotion, tu2024stableanimator, wang2024unianimate, huang2024makeyouranchor} to DiT-based methods~\cite{shao2024human4dit, wang2025unianimatedit},
which progressively improves motion fidelity and temporal coherence.
For audio-driven control, representative works from head animation~\cite{zhang2023sadtalker, ji2024sonic, tian2024emo, chen2025echomimic}, semi-body animation~\cite{corona2024vlogger, meng2024echomimicv2,lin2024cyberhost} to full-body animation~\cite{lin2025omnihuman}.
However, these methods primarily focus on single-subject animation. Even as model capacities improve, enabling plausible motion generation from images depicting well-structured human interactions, the lack of attention to non-human subjects (e.g., objects, environments) still constrains the ability of current systems to achieve controllable human-object interaction video generation.

\subsection{Human-object Interaction Video Generation}
As controllable human video generation continues to improve, a key emerging agenda is how to enable human-object interaction grounded on controllable human motion. Outside of video generation, researchers have focused on generating human-object and human-scene motion~\cite{jiang2024trumans, jiang2024autonomous,xu2023interdiff,li2023object,peng2023hoi,ghosh2023imos}. Within video generation, some approaches aim to retain real human-object interactions in existing videos while replacing either the human or the object. For example, MIMO~\cite{men2024mimo} and AnimateAnyone2~\cite{hu2025animateanyone2} attempt to substitute human subjects in complex dynamic interaction scenes, whereas HOI-Swap~\cite{xue2024hoi} and ReHoLD~\cite{fan2025rehold} focus on replacing handheld objects within a real video. However, these approaches rely on real videos as a foundation and are thus limited to editing existing content, lacking the ability to generate human-object interactions from scratch.

Other works have explored object injection in human animation to human-object interaction video generation from scratch. 
For instance, AnchorCrafter~\cite{xu2024anchorcrafter} fine-tunes target objects and utilizes multi-conditional inputs such as pose and depth to represent interaction-aware human motion. ManiVideo~\cite{pang2024manivideo} achieves accurate control of interaction motion through precise 3D modeling of both the human and the object. 
Nevertheless, these methods rely on strong conditional inputs, making them difficult to apply in unconstrained scenarios.
Our proposed HunyuanVideo-HOMA explores a weak HOI condition for user-friendly generation and accomplishes high generalizability of the objects.
\section{Method}
\label{sec:method}
In this section, we propose the weak condition for HOI video generation in Sec.~\ref{sec:method.weakcondition}, our weak condition driven network architecture in Sec.~\ref{sec:method.architecture}, and the training strategies and data curation in Sec.~\ref{sec:method.training}.

\subsection{Weak HOI Condition Setting}
\label{sec:method.weakcondition}

\begin{figure}[htbp]
    \centering
    \includegraphics[width=\linewidth]{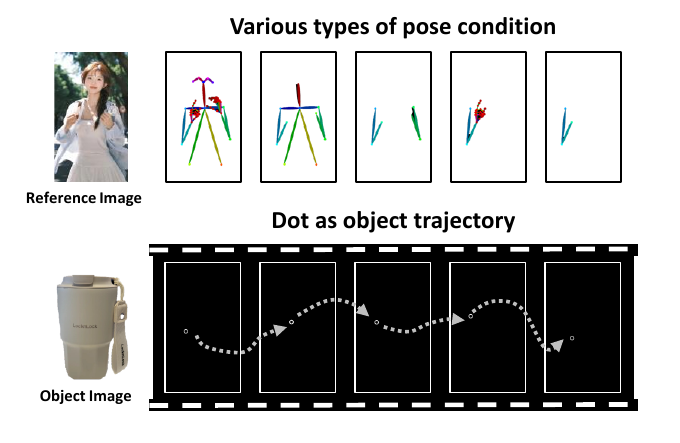}
    \caption{We explore a weak HOI condition, where each component of the human pose can be optionally removed, and the object’s motion is a sequence of dots indicating the center points of its trajectory.}
    \label{fig:method.weakconditions}
\end{figure}
A core challenge in HOI video generation is acquiring expressive yet editable motion inputs. Existing methods rely on actor performances or just swapping objects, limiting scalability and user controllability. We propose a sparse, user-editable input framework that leverages minimal but semantically meaningful conditions to generate realistic, physically plausible interactions. 

We formulate weakly conditioned HOI video generation as animating a reference human image $I_H$ with an inserted object image $I_O$, producing an output video $Y = \{y^{1:n}\}$ with human-object interaction of $n$ frames. Two separate motion sequences guide the human-object interactions under intentionally weak controls (Fig.~\ref{fig:method.weakconditions}). Human motion is driven by sparse skeletal sequences $P = \{p^{1:n}\}$, selectively removing all body parts except the arms in contact with the object; finger skeletons are optionally excluded to reduce difficulty for users to specify, with finger motion inferred contextually. Object trajectory is represented by a dot sequence $D = \{d^{1:n}\}$ indicating the object’s center point, enabling autonomous orientation inference and avoiding the unnatural constraints of naive bounding-box methods.

Our framework also supports text $T$ and audio $A = \{a^{1:n}\}$ as complementary control signals. The text offers intuitive semantic guidance, while audio enables the animated human not only to act but also to talk as expected in an end-to-end manner.

\subsection{Architecture}
\label{sec:method.architecture}

\subsubsection{Preliminaries}
We propose our framework to accomplish the weak-conditioned HOI video generation.
Our network architecture is modified from HunyuanVideo-T2V~\cite{kong2024hunyuanvideo}, including an MMDiT-based~\cite{esser2024sd3-mmdit} video diffusion model and a 3D VAE model~\cite{kingma2013vae} with encoder $Enc$ and decoder $Dec$.
During training, the video sequence $X$ is encoded into the latent space as $Z_0=Enc(X)=\{z_0^{1:n}\}$ and flow matching~\cite{lipman2022flowmatching} is used as the training objective.
During inference, a set of initial latent variables $U=\{u^{1:n}\}$ is sampled from standard Gaussian noise and progressively denoised over $T$ timesteps to produce an estimate $\hat{Z_0}$. The final output video is then obtained via decoding, $Y=Dec(\hat{Z_0})=\{y^{1:n}\}$.

\subsubsection{Context Fusion}
\begin{figure*}
    \centering
    \includegraphics[width=\linewidth]{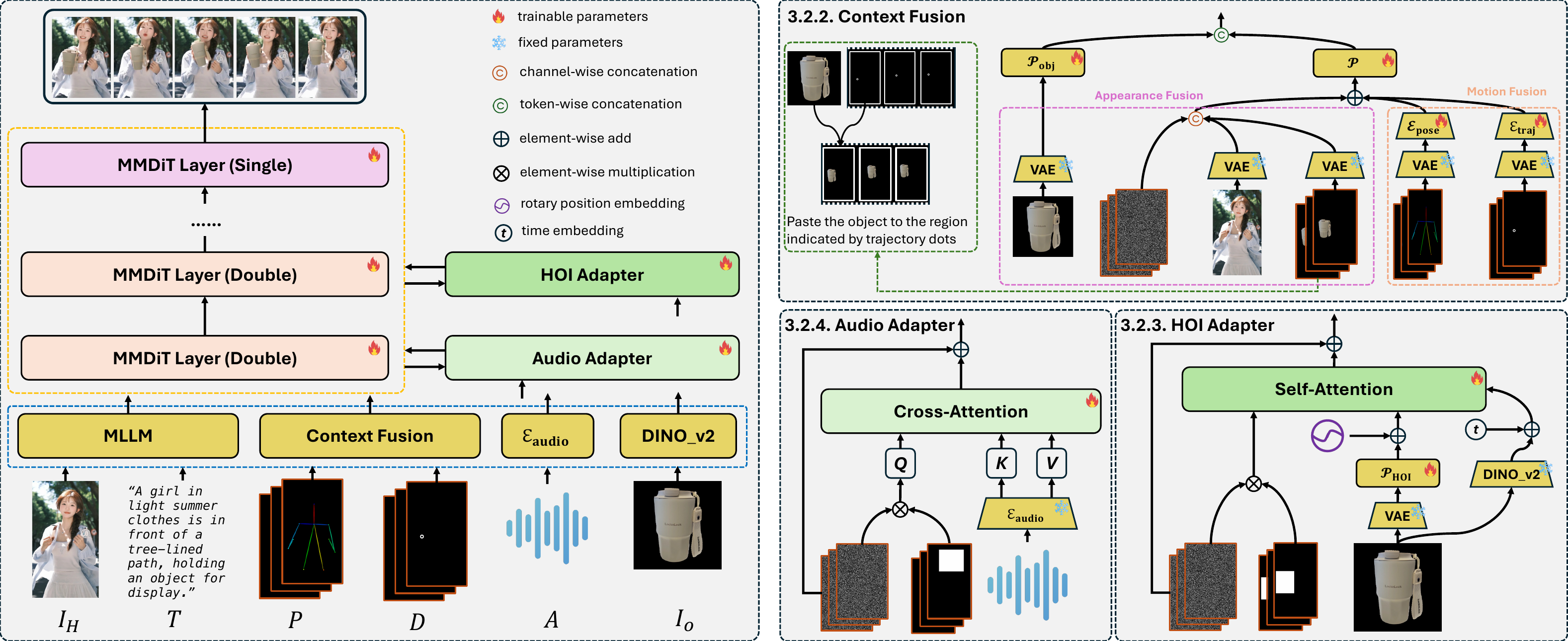}
    \caption{\textbf{Architecture of HunyuanVideo-HOMA. HunyuanVideo-HOMA is built upon the multi-modal diffusion transformer and is composed of context fusion, HOI adapter, and audio adapter.}}
    \label{fig:method.architecture}
\end{figure*}

To effectively leverage the model’s generative prior, we propose a context fusion strategy for object and human injection in the input spaces. The overall architecture is illustrated in Fig.~\ref{fig:method.architecture}. Context fusion involves two spaces, namely, the latent space and token space. The latent space is encoded by the VAE encoder as $Z=Enc(X)$, RoPE~\cite{su2024roformer-rope} is added to provide position information. Subsequently, a patch embedding layer $\mathcal{P}$ maps the latent features into the token space, yielding $H = \mathcal{P}(Z)$. 
In the following, we introduce our context fusion strategy from two perspectives: appearance and motion.

\paragraph{Appearance Context Fusion}
The main purpose of this module is to maintain the appearance of human image $I_H$ and injected object $I_O$ during animation. Due to the different requirements of humans and objects, we propose a separate manner to inject their appearance.
For the human image $I_H$ to be animated, we propose a channel-concat style injection approach performed on input latent space:
\begin{align}
    Z_{cat} = concat_c(Z, Z_{ref}),
\end{align}
where $Z$ is the input noise, $Z_{ref}$ is the VAE feature extracted from $I_H$, $Z,Z_{ref}\in[b,f,h,w,c]$, $Z_{cat}\in[b,f,h,w,3c]$, and $concat_c$ is the channel-wise concat operation. The output embedding $Z_{cat}$ is then input into the DiT blocks. The proposed channel concatenation scheme aims to maintain the structural and semantic content of the reference image.
As for the object image $I_O$, we perform object injection from both the token space and the latent space. 
In token space, we adopt a temporal-concat style scheme:
\begin{align}
    H = concat_f(H_{obj}, H),
\end{align}
where $H_{obj}\in[b,1,h,w,c]$ is the object feature extracted from $I_O$ as $H_{obj}= \mathcal{P}(Enc(I_O))$ and $concat_f$ is the temporal-wise concat operation. After temporal-concat, the output $H$ is in $[b, f+1, h, w, c]$. The RoPE is set with offset=-1 for the temporal-concat image.
Unlike reference human images that aim to preserve original content, object appearances are inserted into the video content. Therefore, we adopt an injection strategy that aligns more closely with the video generation context—specifically, the object is injected at frame -1 and emerges in the subsequent frames. In addition, given the effective injection of the human representation in the latent space, we introduce object appearance features into the latent space to further improve stability. The object is copy-pasted to the image region indicated by trajectory to serve as strong appearance prior:
\begin{equation}
    Z_{obj|D} = \{Z^{i:n}_{obj}| Z^i_{obj} = Z_{obj}|_{D^i} \},
\end{equation}
and the features are then concatenated along the channel dimension in the latent space. The final channel concatenation is:
\begin{equation}
    Z_{cat} = concat_c(Z, Z_{ref}, Z_{obj|D}),
\end{equation}
where the final dimension of $Z_{cat}$ is in $[b,f,h,w,3c]$.
Beyond improving injection consistency, the size of spatially aligned and pasted object $Z_{obj|D}$ also affects its final scale in generated video. Importantly, ours does not suffer from noticeable copy-paste artifacts.

To enhance global semantics, we extract LLaVA~\cite{liu2023llava} visual features from both human and object images and fuse them with the feature of text $T$, and feed it to MMDiT’s text branch.

\paragraph{Motion Context Fusion}
\label{sec:method.arch.motion}
We perform motion context fusion in latent space as well.
For the injection of human pose and object trajectory, we also adopt a separate-then-merge style module, where the two motion conditions are processed independently before being fused. For the pose sequence $P$ and the object trajectory sequence $D$, we first use the 3D VAE to map them into feature representations $Z_{pose}=Enc(P)$ and $Z_{obj}=Enc(D)$, and then use separate one-layer CNN $\mathcal{E}_{pose}$ and $\mathcal{E}_{traj}$ to project them into latent space as $Z_{pose} = \mathcal{E}_{pose}(Z_{pose})$ and $Z_{obj} = \mathcal{E}_{traj}(Z_{obj})$.
Human pose and object trajectory features are then added to the appearance channel-concat latent to get the final input:
\begin{equation}
    Z = Z_{cat} + Z_{pose} + Z_{traj}.
\end{equation}
We observe that the separate-then-merge design leads to better appearance preservation. The analysis is provided in the ablation study.

\subsubsection{HOI adapter}
We observed that the model struggled to learn sufficient information about the appearance of the object. We attribute this to weak gradient propagation when injection is limited to the input layer, due to the network’s depth. To address this, we propose the HOI Adapter—a multi-layer injection strategy that introduces object features at several points within the network.
We posit that effective adapter design relies on leveraging the original model’s parameter space, rather than constructing entirely new representational branches. Therefore, we modify the structure of the original MMDiT's double-stream block, where the text token is replaced by object token $H_{obj}$, and the weights are initialized with pretrained parameters. The HOI Adapter is formulated as:
\begin{align}
    H= H +  Mask_{obj} \otimes SelfAttn([H, H_{obj}+RoPE(-2)]), 
\end{align}
where the object feature $H_{obj}$ is token-concat with the hidden feature $H$, followed by a self-attention layer for feature interaction. The adapter output is added back to $H_{noise}$ within the object region $Mask_{obj}$ to provide a precise location. We also use a DINO~\cite{caron2021dinov1} feature of $I_{O}$ combined with time embedding to provide semantic information.
By injecting object information across multiple layers, the model learns object injection more effectively and accurately. To preserve compatibility with the original representation space, we initialize the adapter's self-attention weights from the corresponding MMDiT blocks. HOI Adapters are inserted only into the even-numbered layers.

\subsubsection{Audio Cross-attention Adapter}
\label{sec:audio_adapter}
We propose a cross-attention style injection module for audio input. The input audio feature $A$ is mapped with an MLP layer into $H_{audio}$, then we utilize a cross-attention layer to modify the face region:
\begin{align}
    H = H + Mask_{face} \otimes Softmax(\frac{Q_H K_{A}^T}{\sqrt{d}})\cdot V_A,
\end{align}
where $Q_H=W_{QH} \cdot H$, $K_A = W_{KA} \cdot H_A$, $V_A = W_{VA} \cdot H_A$, and $Mask_{face}$ is the mask with the face-region.
The cross-attention adapters are added to the even-numbered layers of MMDiT blocks. 

\subsection{Training Strategy and Data Curation}
\label{sec:method.training}
We adopt a three-stage training strategy. Stage one focuses on learning human pose control using human-only datasets, with appearance injected via channel-concatenated reference images and a pose encoder $\mathcal{E}_{pose}$. In stage two, we introduce HOI data at 512$\times$512 resolution, enabling the model to learn object control while incorporating pose and audio conditions. $\mathcal{E}_{traj}$ is initialized from $\mathcal{E}_{pose}$. Stage three increases the resolution to 512$\times$896 to enhance visual quality and high-resolution generation capability.

To address the scarcity of HOI training data and annotations, we propose an automated pipeline for selecting human-object interaction (HOI) video clips, comprising three stages:
\begin{itemize}[leftmargin=10pt]
\item Step-\uppercase\expandafter{\romannumeral1}: HOI Recognition. We prompt Qwen-VL~\cite{wang2024qwen2-vl} to identify whether a video contains HOI and caption objects.

\item Step-\uppercase\expandafter{\romannumeral2}:  HOI Extraction. Using the generated object captions, we apply Grounding-DINO~\cite{liu2024grounding} and SAM2~\cite{ravi2024sam2} to perform object detection and segmentation.

\item Step-\uppercase\expandafter{\romannumeral3}: HOI Filtering. To reduce false positives (e.g., irrelevant background objects), we use DepthAnythingV2~\cite{yang2024depthanythingv2} to estimate depth maps. We compare the mean depths of the segmented object and hand regions, retaining only those with similar depths to ensure plausible physical interaction.
\end{itemize}

\begin{figure}[t]
    \centering
    \includegraphics[width=1.0\linewidth]{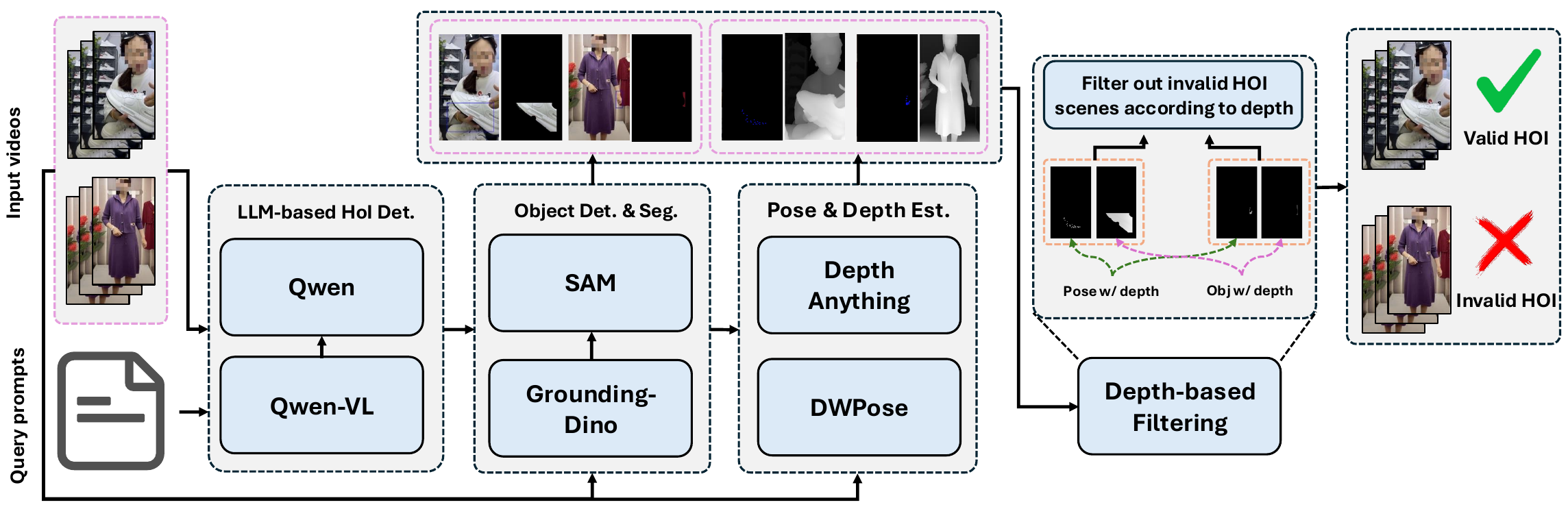}
    \caption{\textbf{Depth-aware HOI Data Curation.}}
    \label{fig:method.data}
\end{figure}
\section{Experiments}

\subsection{Experimental Settings}

\subsubsection{Implementation Details}
We use a fixed learning rate of 1e-5 across all stages. Training is conducted on 48$\times$96G GPUs with a total batch size of twelve. The total training consists of 16,000 steps in stage one, followed by 2,000 steps in stage two, and 5,000 steps in stage three. To improve generalization and robustness, we apply data augmentation to the object images, including random scaling, rotation, and position shifting. Human pose information is extracted using DWPose~\cite{yang2023dwpose}, while audio features (when applicable) are processed using Librosa~\cite{mcfee2015librosa}.
Each generated video has a duration of five seconds.

\subsubsection{Dataset}
We apply depth-aware HOI(Human-Object Interaction) data curation to internet-sourced videos to support training under diverse interaction scenarios and finally collect approximately 140 hours of training data.
For evaluation, we conduct experiments on two test sets:
\textbf{AnchorCraft test set}: The AnchorCrafter~\cite{xu2024anchorcrafter} test set includes four objects and four persons with simple object types and basic interaction motions. We reorganize the AnchorCraft test set via resampling, resulting in 80 clips. 
\textbf{Self-collected test set}: To evaluate more challenging scenarios, we construct a custom test set from Internet videos, containing 100 clips with complex human-object interactions and diverse object appearances.
Audio inputs are randomly collected for both test sets. 
We conduct quantitative and qualitative evaluations on each test set to assess the model's generalization ability across varying levels of interaction difficulty.
\textbf{Due to page limit, AnchorCrafter test set results are in the appendix.}

\subsubsection{Comparison Methods}
We compare our method with the following methods:
AnchorCrafter~\cite{xu2024anchorcrafter},
VACE-14B~\cite{jiang2025vace},
MimicMotion~\cite{zhang2024mimicmotion},
StableAnimator~\cite{tu2024stableanimator},
UniAnimate-Dit~\cite{wang2025unianimatedit}, and
EchoMimic-v2~\cite{meng2024echomimicv2}. 
AnchorCrafter~\cite{xu2024anchorcrafter} is a human-object interaction video generation approach that requires fine-tuning on the target object.
VACE-14B~\cite{jiang2025vace} is a general video editing framework based on the 14B video model Wan~\cite{wang2025wan} and we use the pose control setting. MimicMotion~\cite{zhang2024mimicmotion}, and UniAnimate-DiT~\cite{wang2025unianimatedit} generate motion from a person image and a sequence of poses. EchoMimic-v2~\cite{meng2024echomimicv2} takes audio and a human image as input. 
For methods not originally designed for human-object interaction, we composite the object into the person image to approximate an interactive scenario. 
For a fair comparison, all compared methods utilize the original motion inputs, and our method uses the whole full pose for inference.
AnchorCrafter requires finetuning on the target object and is compared only in the AnchorCrafter test set.

\subsubsection{Metrics}
We evaluate our method using several established metrics. For visual quality, we use
FID~\cite{heusel2017fid} and 
FVD~\cite{unterthiner2018fvd} to evaluate the image and video quality.
Object CLIP (OC)~\cite{xu2024anchorcrafter} is used to evaluate the object consistency by computing the CLIP~\cite{radford2021clip} similarity of the objects between ground-truth and generated videos within the bounding box.
Additionally,
Subject Consistency (SC)~\cite{xue2024hoi},
Background Consistency (BC)~\cite{xue2024hoi},
and Motion Smoothness (MS)~\cite{xue2024hoi} are employed to measure general video quality.
For hand motion quality, we calculate 
Hand Agreement Score (HAS)~\cite{xue2024hoi} and
Hand fidelity (HF)~\cite{xue2024hoi}. 
To evaluate lip synchronization, we adopt the commonly used 
Sync-C~\cite{syncnet}, which measures the alignment confidence between visual and audio content.

\subsection{Comparison Results}
\subsubsection{Quantitative Results}
\label{expr:main_quan}
As shown in Table~\ref{tab:main}, our method achieves the best FID and FVD scores, indicating superior visual quality. The Object CLIP (OC) score is notably the highest among all methods. This highlights the strong ability of our framework to preserve object appearance. The results on SC, BC, and MS are comparable to others, reflecting solid structural coherence and smooth motion. Moreover, on our self-collected test set with more complex interaction scenarios, we achieve the highest Hand Agreement Score (HAS), indicating strong generalization under challenging conditions. Given the limitations of objective metrics in evaluating fine-grained hand-object interactions, the user study provides a more comprehensive and trustworthy assessment of interaction quality. Sync-C scores are also comparable to Echomimic-v2, validating the effectiveness of our audio injection design.

\subsubsection{Qualitative Results}

\begin{figure*}[htbp]
    \centering
    \includegraphics[width=\linewidth]{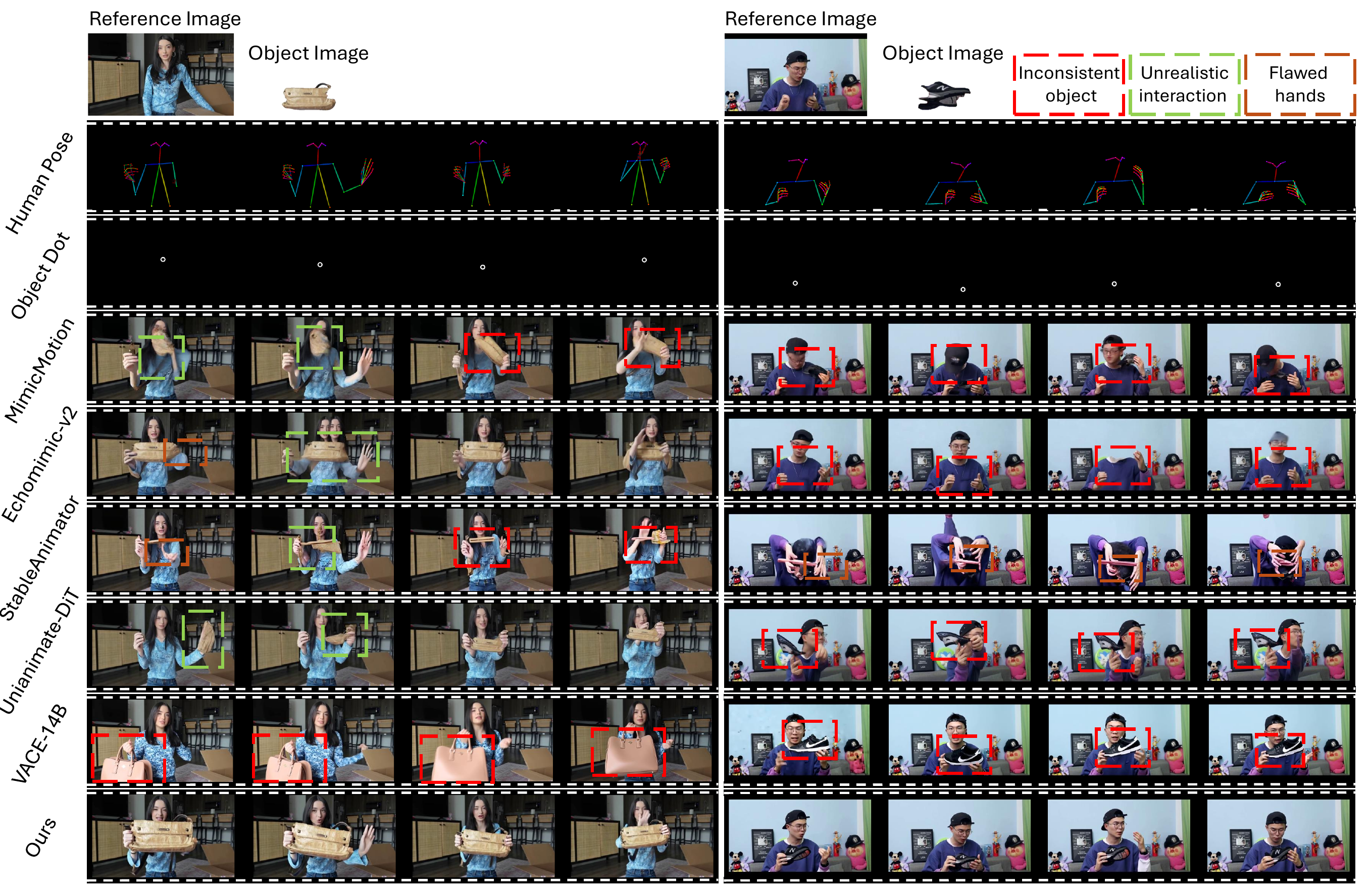}
    \caption{Comparison with SOTAs on self-collected test set. More results can be found in the supplementary materials.}
    \label{fig:expr.main_expr_selfcollect}
\end{figure*}
In Fig.~\ref{fig:expr.main_expr_selfcollect}, we present the qualitative comparisons of the self-collected test set, where our method achieves better preservation of human and object appearances, along with more consistent and visually stable results. MimicMotion produces reasonably accurate hand poses but fails to generate the object. Echomimic-v2 suffers from inconsistent object generation and noticeable flickering artifacts on the character, while StableAnimator tends to produce flawed or anatomically incorrect hands. Unianimate-DiT maintains good subject consistency but struggles to generate realistic and coherent interaction motions. VACE-14B produces high-quality videos overall but fails to faithfully preserve the appearance of the object. In contrast, our method generates high-quality human-object interaction videos that are visually coherent, temporally stable, and faithful to the reference inputs.
\begin{table}
  \centering
  \caption{Quantitative results of our method compared with SOTAs. }
    \label{tab:main}
  \resizebox{1\linewidth}{!}{
  \begin{tabular}{@{}lccccccccc@{}} 
    \toprule
      \multirow{2}{*}{Method} & \multicolumn{9}{c}{\textbf{Self-collected test set}} \\
                                                              & FID↓            & FVD↓            & OC↑           &SC↑           &BC↑             & HAS↑                & HF↑                   & MS↑           & Sync-C↑ \\
    \midrule
    MimicMotion                                               &  83.56          & 1072.02         & 81.70         &  93.48       & 93.82         & 87.51               &  99.16                &99.08            &  -\\
    Echomimic-v2                                              &  100.20         & 1332.06         & 79.82         &  94.65       & 92.92         & 86.84               & 99.54                 &99.04            &  4.12\\
    StableAnimator                                            &  125.84         & 1530.48         & 79.26         &  88.19       & 91.36         &  83.98               & 94.97                &98.42            &  -\\
    UniAnimate-DiT                                            &  83.18          & 844.32          & 83.74         &\textbf{95.82}& \textbf{94.96}& 83.17               & 99.48                 &\textbf{99.27}   & -\\
    VACE-14B                                                  &  77.07          & 968.79          & 82.88         &  94.73       & 95.59         & 75.74                & \textbf{99.87}       &99.04            &  -\\
    Ours                                                      &  \textbf{51.60} & \textbf{502.69} & \textbf{90.05}&  95.19       & 94.45         & \textbf{87.93 }      & 98.76                &99.14            &   \textbf{4.33}   \\
    \bottomrule
  \end{tabular}
  }
\end{table}

\subsection{Ablation Studies and Discussion}
\label{sec:expr.ablation}
To facilitate efficient comparison across different variants, all ablation experiments are conducted at a resolution of 512$\times$512 with 2,000 training steps. The self-collected test set is utilized to verify the ablations studies. We focus on key components of our framework, including the object motion setting and architecture variants. 
\begin{table}
  \centering
  \caption{Ablation results with 512$\times$512 resolutions.}
  \label{tab:ablation}
  \resizebox{1\linewidth}{!}{
  \begin{tabular}{@{}lcccccccc@{}}
    \toprule
   Method                             & FID↓            & FVD↓           & OC↑           &SC↑          &BC↑           & HAS↑     & HF↑    & MS↑         \\
    \midrule
    Ours                              &  38.32          & 386.83         & 88.75         & 93.87       & 93.65        & 88.23         & 98.70       &  98.78       \\
   \midrule
    w/ bbox                           &  76.73          & 720.24         & 82.23         & 92.86       & 93.86        & 85.73         & 98.36       &  98.89       \\
    w/ gaussian dot                   &  86.54          & 691.56         & 80.74         & 94.46       & 94.60        & 85.69         & 98.88       &  99.05       \\
    \midrule
    w/o  token concat                 &  74.38          & 681.18         & 82.18         & 93.88       & 93.86        & 84.73         & 98.34       &  98.64       \\
    w/o channel concant               &  74.38          & 799.24         & 84.90         & 94.04       & 94.14        & 85.29         & 97.30       &  98.93       \\
    w/ fix-copy                      &  58.48          & 588.63         & 85.88         & 93.99       & 94.23        & 85.41         & 97.96       &  98.67       \\
   \midrule
    single motion enc.                &  50.34          & 505.91         & 86.82         & 94.31       & 94.42        & 85.62         & 98.86       &  98.99       \\
    \midrule
    w/o adapter                       &  62.09          & 582.38         & 82.01         & 92.04       & 93.60        &  84.87        & 99.48       &  98.90       \\
    w/ cross-attention                 &  52.65          & 592.44         & 83.39         & 92.22            & 93.77             & 85.58         & 98.59       & 98.97             \\
    \bottomrule
  \end{tabular}
  }
\end{table}

\subsubsection{Analysis of Object Motion Setting}
\begin{figure}[p]
    \centering
    \includegraphics[width=1.0\linewidth]{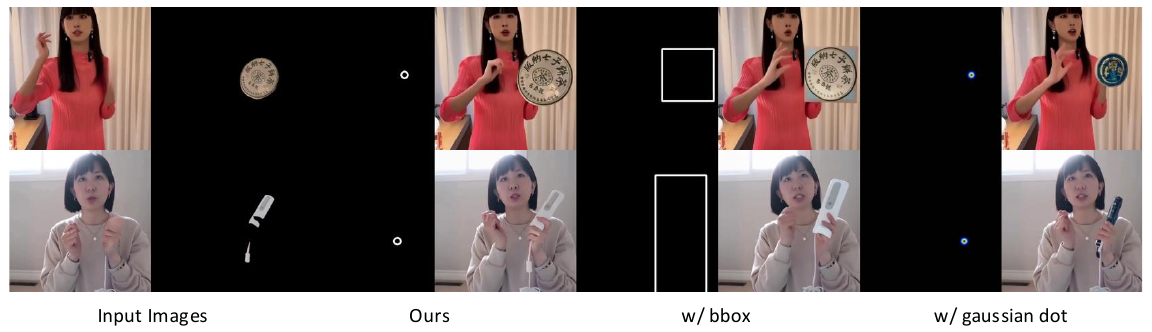}
    \caption{Ablation results of different object motion.}
    \label{fig:method.ablation_objmotion}
\end{figure}

To evaluate our dot-based object motion strategy, we compare three variants: bounding box (bbox), Gaussian-colored dot (Gaussian dot), and the proposed dot trajectory (dot). As shown in Fig.\ref{fig:method.ablation_objmotion} and Table\ref{tab:ablation}, bbox control enforces explicit constraints on position, size, and orientation, but often results in rigid or unnatural motion during complex interactions. Gaussian dot leads to incorrect object appearance due to distribution mismatch with the pre-trained motion encoder, impeding convergence. In contrast, our dot trajectory control yields smoother and more natural trajectories by imposing minimal spatial constraints, enabling better adaptation to human motion dynamics.

\subsubsection{Analysis of Context Fusion}
\begin{figure}[p]
    \centering
    \includegraphics[width=1.0\linewidth]{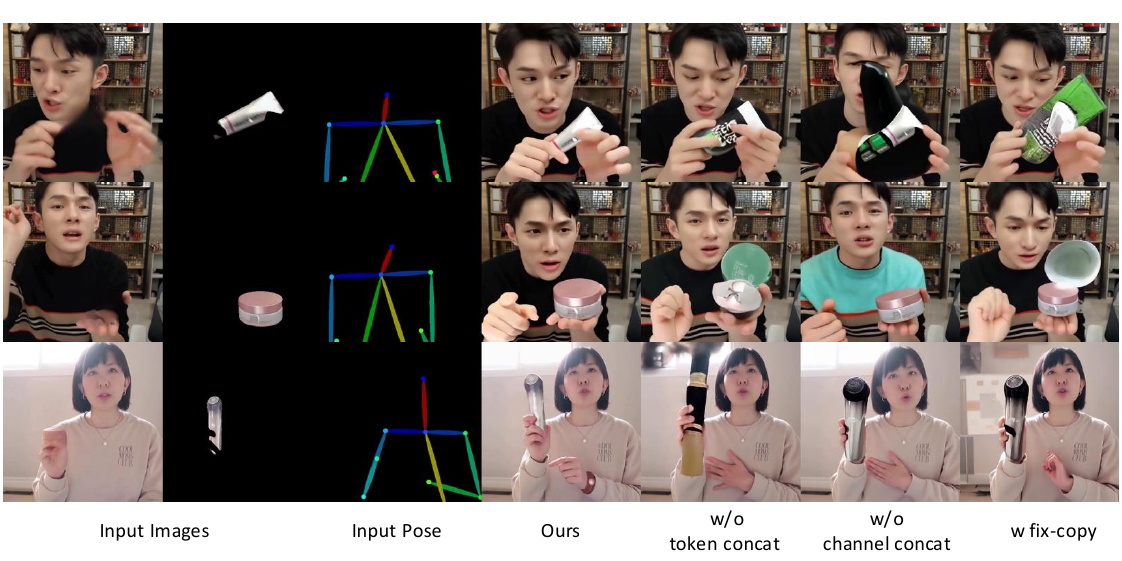}
    \caption{Ablation results of context fusion.}
    \label{fig:method.ablation_contextfusion}
\end{figure}

We evaluate the impact of appearance and motion context in our network via ablation. For object appearance context, we compare three variants: (1) without token concat, (2) without channel concat, and (3) with fixed-location channel concat (w/ fix-copy). Results (Table~\ref{tab:ablation}, Fig.~\ref{fig:method.ablation_contextfusion}) show that all variants reduce appearance fidelity. Removing the token concat has the most significant effect, highlighting its importance for conveying fine-grained textures. Retaining token concatenation while removing channel concatenation preserves partial appearance, though alignment issues with pose motion lead to degraded coherence. Fixing channel concat to a static location is less effective than our alignment strategy, which avoids copy-paste artifacts and ensures semantically meaningful guidance.

\begin{figure}[p]
    \centering
    \includegraphics[width=1.0\linewidth]{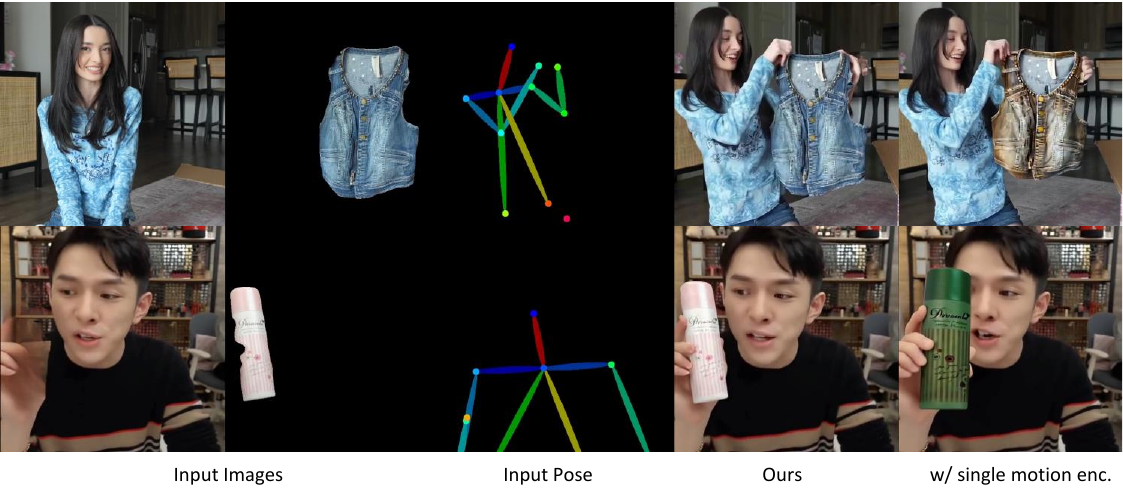}
    \caption{Ablation results of motion encoder.The qualitative results indicate that employing the separate motion encoder enables more effective disentanglement between the appearances of the human and the object.}
    \label{fig:method.ablation_motionenc}
\end{figure}

For motion context fusion, we evaluate an alternative encoding strategy (Fig.\ref{fig:method.ablation_motionenc}, Table\ref{tab:ablation}) in which human pose and object motion are fused into a single composite motion map and processed by a shared motion encoder. This naive fusion leads to a noticeable drop in object appearance consistency. We attribute this to motion entanglement: merging heterogeneous motion sources without modality separation introduces semantic interference, degrading the model’s ability to encode and preserve object-related motion effectively.

\subsubsection{Analysis of HOI adapter}
\begin{figure}[p]
    \centering
    \includegraphics[width=1.0\linewidth]{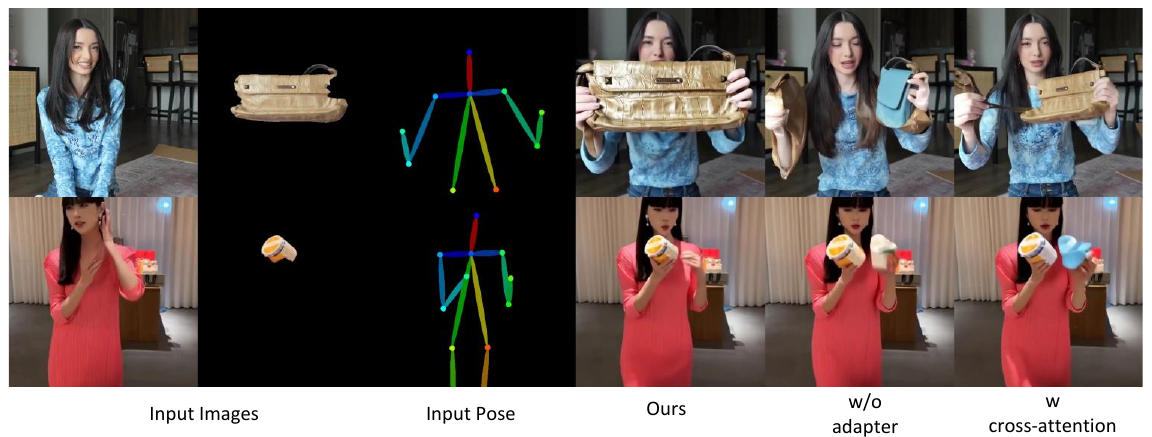}
    \caption{Ablation results of HOI adapter.
    Without HOI adapter, the capability to track object motion is not well disentangled; instead, it is inherently coupled with the human pose.}
    \label{fig:method.ablation_adapter}
\end{figure}

We assess the effectiveness of our proposed HOI Adapter by comparing two variants: (1) removing the adapter entirely, and (2) replacing its self-attention with a cross-attention layer akin to the audio adapter (Sec.\ref{sec:audio_adapter}). As shown in Table\ref{tab:ablation} and Fig.~\ref{fig:method.ablation_adapter}, removing the adapter leads to significantly slower convergence and failure to follow the motion dot, often resulting in the object appearing on the incorrect hand. The cross-attention variant also suffers from poor appearance preservation and inaccurate position injection, likely due to distribution mismatch. In contrast, reusing the pretrained MMDiT self-attention weights aligns more closely with the model’s internal priors, enabling more accurate and efficient HOI injection.

\subsection{User Study}
We conduct a user study to evaluate the quality of generated videos using five subjective metrics: Human consistency, object consistency, motion consistency, interaction naturalness, and video quality. The two test sets are applied in the user study. For each test case, we present the results from all methods and ask 15 users to score them from one to five. The final scores are then computed by averaging the scores given by the users. As shown in Table~\ref{tab:example_userstudy}, our method consistently outperformed others across all criteria.
\begin{table}[!h]
  \centering
  \caption{User study scores.}
  \label{tab:example_userstudy}
  \resizebox{1\linewidth}{!}{
  \begin{tabular}{@{}lccccc@{}}
    \toprule
                                                     &\multicolumn{5}{c}{\textbf{AnchorCrafter test set} / \textbf{Self-collected test set}}                      \\
     \multirow{2}{*}{Method}                         & Human                      & Object                      & Motion                        & Interaction                   &Video    \\
                                                     & Consistency $\uparrow$     & Consistency $\uparrow$      & Consistency $\uparrow$        & Naturalness $\uparrow$        &Quality $\uparrow$         \\
    \midrule
    AnchorCrafter                                    & 3.56/-                     & 3.46/-                      & 2.79/-                        &3.38/-                         &3.27/-                                 \\
    MimicMotion                                      &1.43/1.31                   &  1.33/1.26                  & 1.37/1.32                     & 1.34/1.29                     &1.81/1.47              \\
    StableAnimator                                   &1.55/1.78                   &  1.24/1.18                  & 1.17/1.16                     & 1.31/1.15                     &1.76/1.38          \\
    Echomimic-v2                                     &2.16/1.60                   &  1.25/1.23                  & 1.25/1.33                     & 1.28/1.24                     &1.87/1.48          \\
    UniAnimate-DiT                                   &3.05/2.40                   &  2.64/2.05                  & 2.09/2.14                     & 2.32/2.14                     &2.64/2.20          \\
    VACE-14B                                         &3.23/2.92                   &2.95/2.77                    &2.55/3.09                      &2.92/3.48                      &2.92/3.37                           \\
    Ours                                             &\textbf{3.90}/\textbf{3.52} & \textbf{3.69}/\textbf{3.66} & \textbf{3.15}/\textbf{3.84}   & \textbf{3.61}/\textbf{4.00}   & \textbf{3.59}/\textbf{3.78}     \\
    \bottomrule
  \end{tabular}}
  \vspace{-10px}
\end{table}

\subsection{More Results}
We provide more results to showcase the practical applications and generalization capability of our method.
To avoid copyright issues, we use the generated images as reference human input.

\begin{figure}[p]
    \centering
    \includegraphics[width=\linewidth]{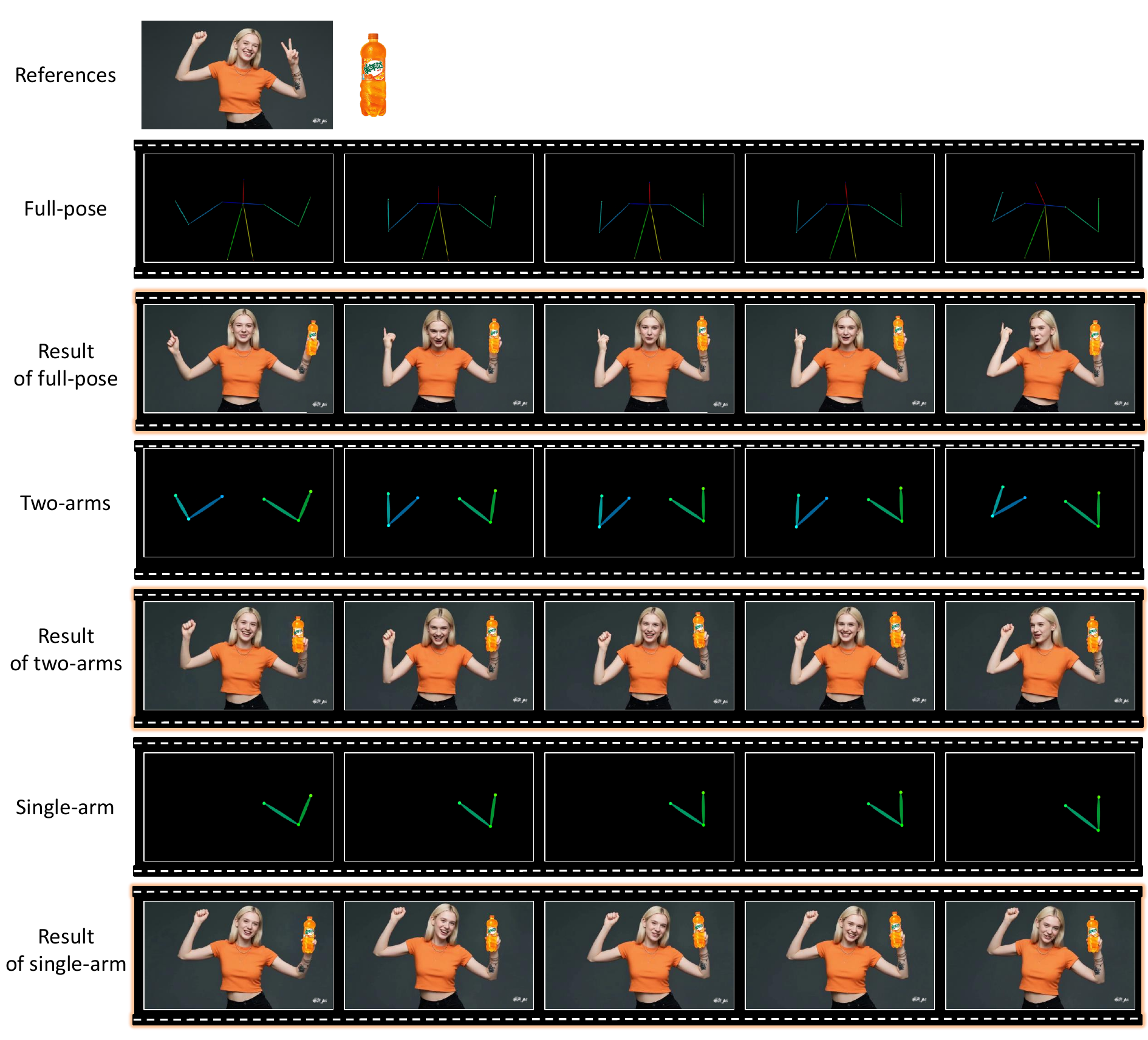}
    \caption{Results with different types of poses.}
    \label{fig:expr.sparsepose}
\end{figure}

\paragraph{Results with Different Pose Types}
We demonstrate the controllability of our method across a variety of weak human pose scenarios, as illustrated in Fig.~\ref{fig:expr.sparsepose}. The results show that our framework effectively adapts to diverse poses while maintaining coherent object interactions and consistent appearance.

\begin{figure}[p]
    \centering
    \includegraphics[width=\linewidth]{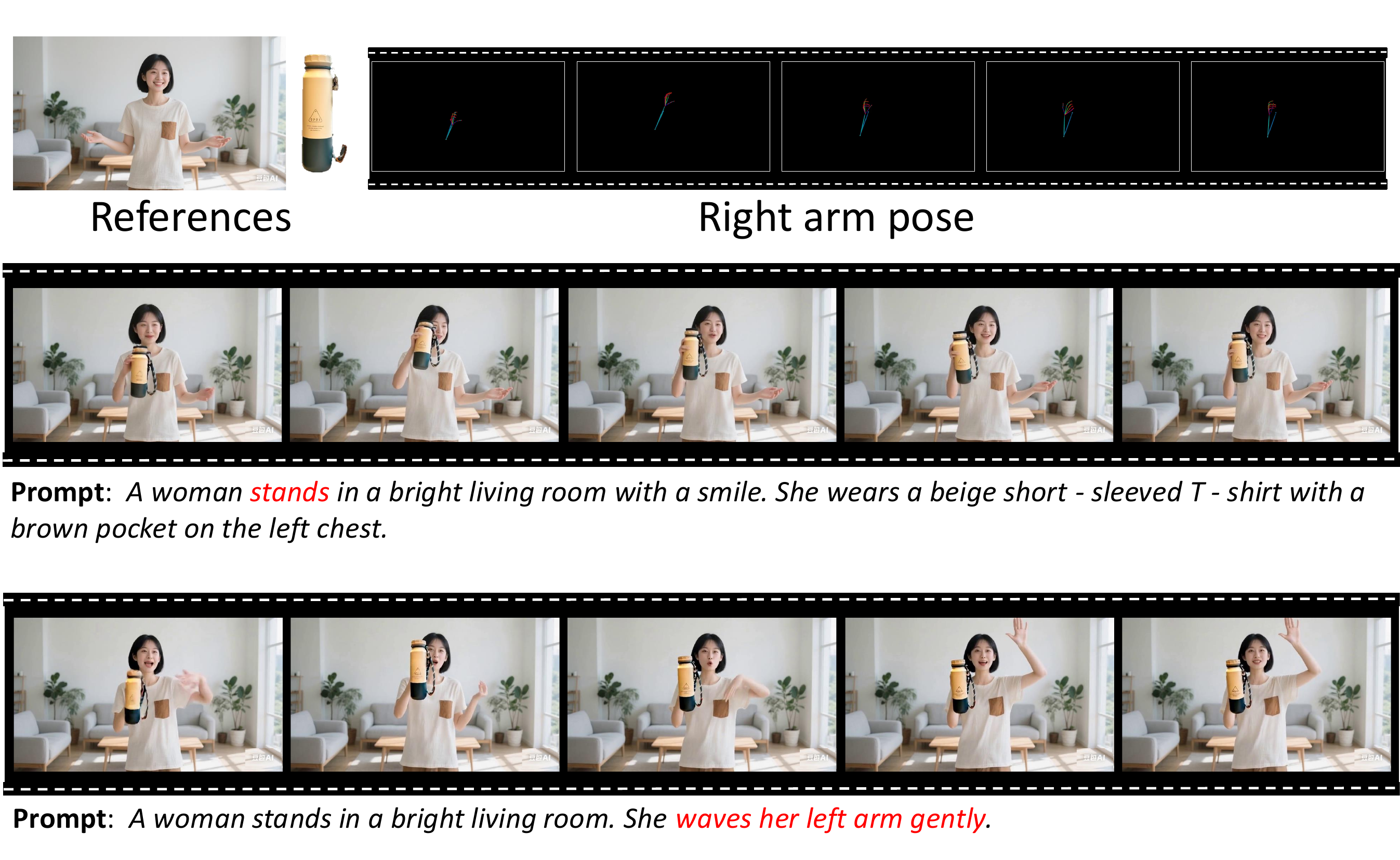}
    \caption{Influence of Text Prompt.}
    \label{fig:expr.textinfluence}
\end{figure}

\paragraph{Influence of Text Prompt}
We explore the impact of combining weak pose guidance with text-based control, as illustrated in Fig.~\ref{fig:expr.textinfluence}. The results demonstrate that our method can effectively integrate minimal pose information with textual semantics to generate coherent and semantically aligned motions.

\begin{figure}[p]
    \centering
    \includegraphics[width=\linewidth]{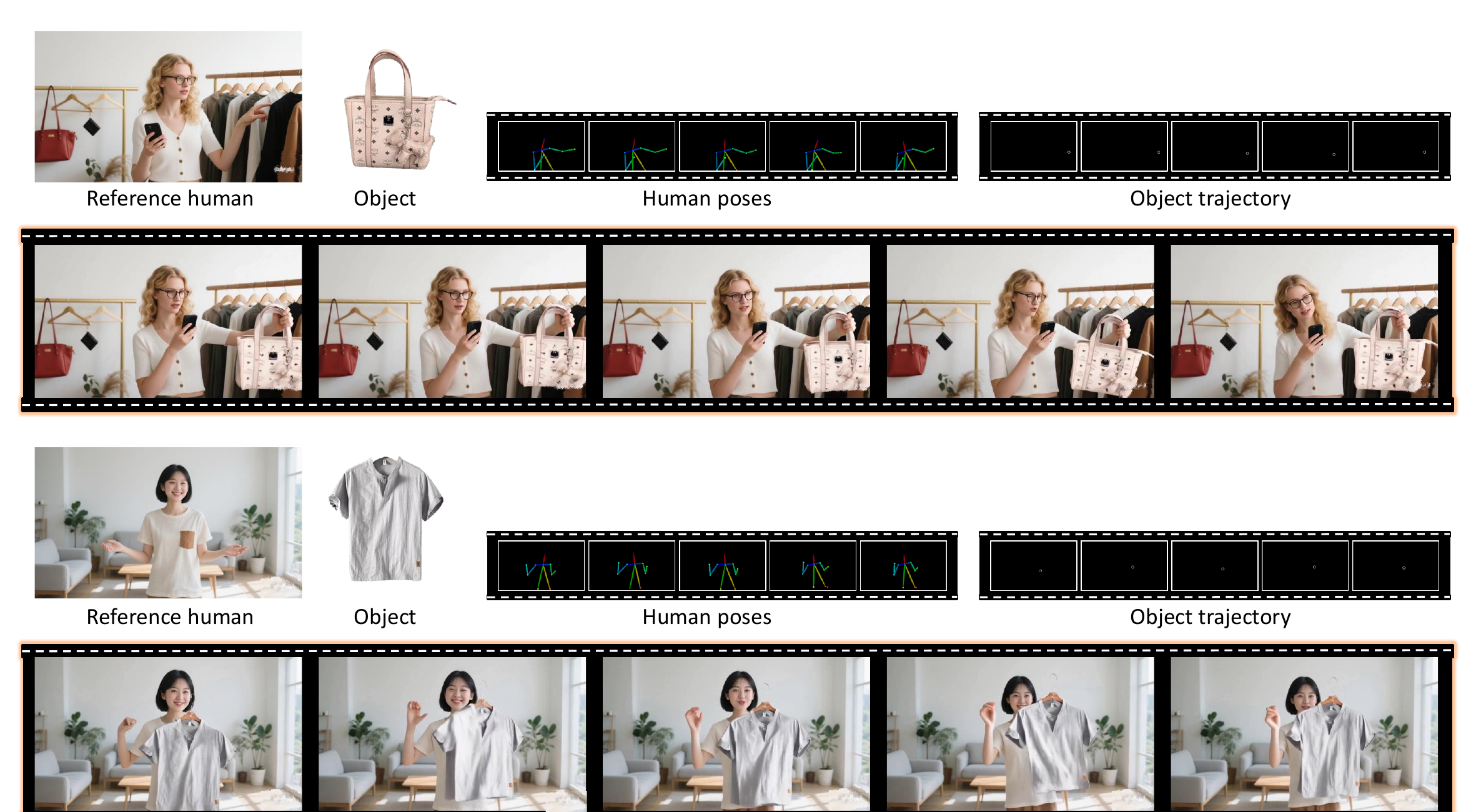}
    \caption{Results of Non-Rigid Objects.}
    \label{fig:expr.flexibleobj}
\end{figure}

\begin{figure*}[p]
    \centering
    \includegraphics[width=\linewidth]{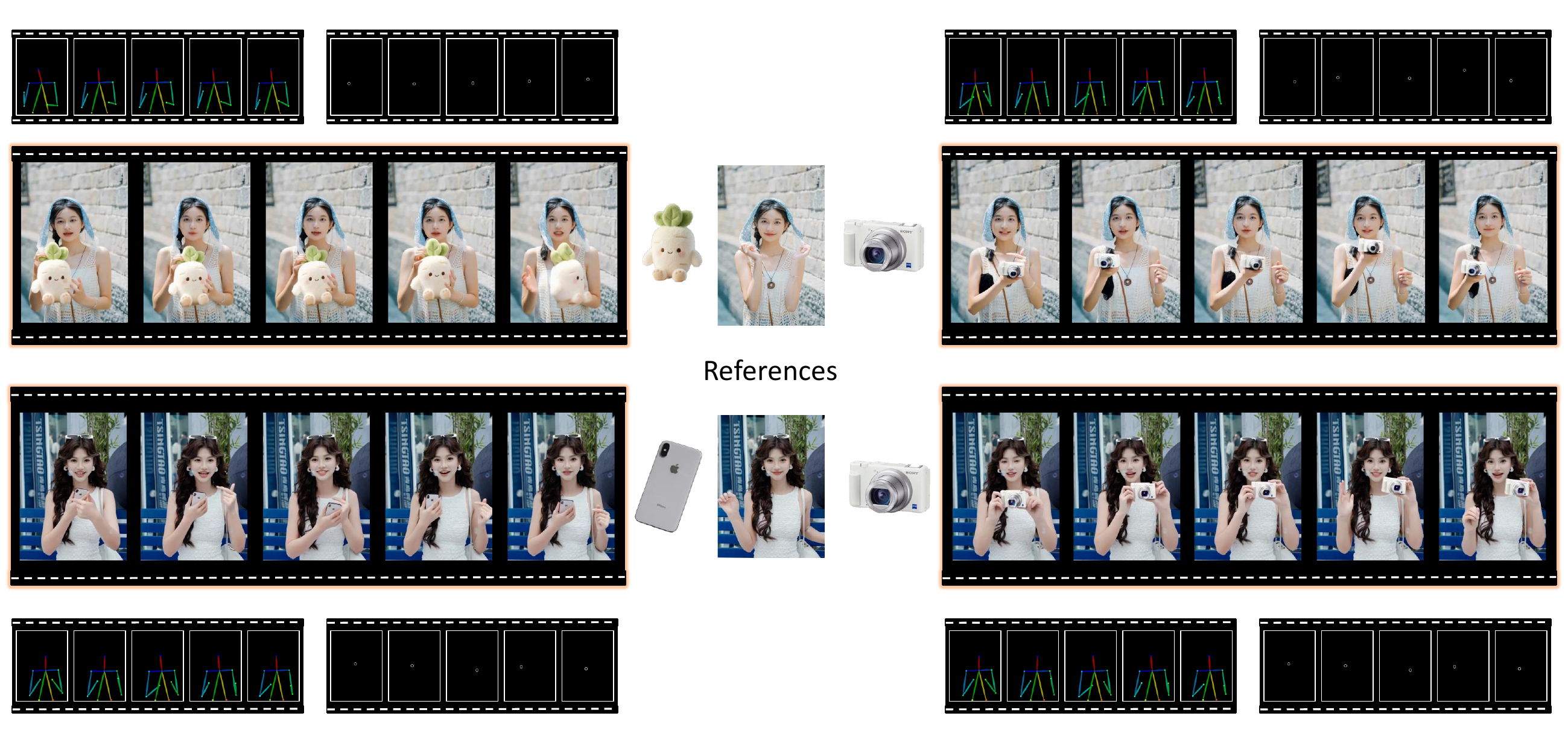}
    \vspace{-20px}
    \caption{Various object appearance and motion.}
    \label{fig:expr.replacements}
\end{figure*}

\paragraph{Variations of Object Appearance and Motion}
We present additional experimental results to demonstrate the versatility of our method. The results of various object appearances and motions are shown in Fig.~\ref{fig:expr.flexibleobj} and Fig.~\ref{fig:expr.replacements}, including flexible objects. The results highlight our model's ability to adapt to varying object appearances while maintaining consistent identity and motion coherence.

\begin{figure*}[p]
    \centering
    \includegraphics[width=0.9\linewidth]{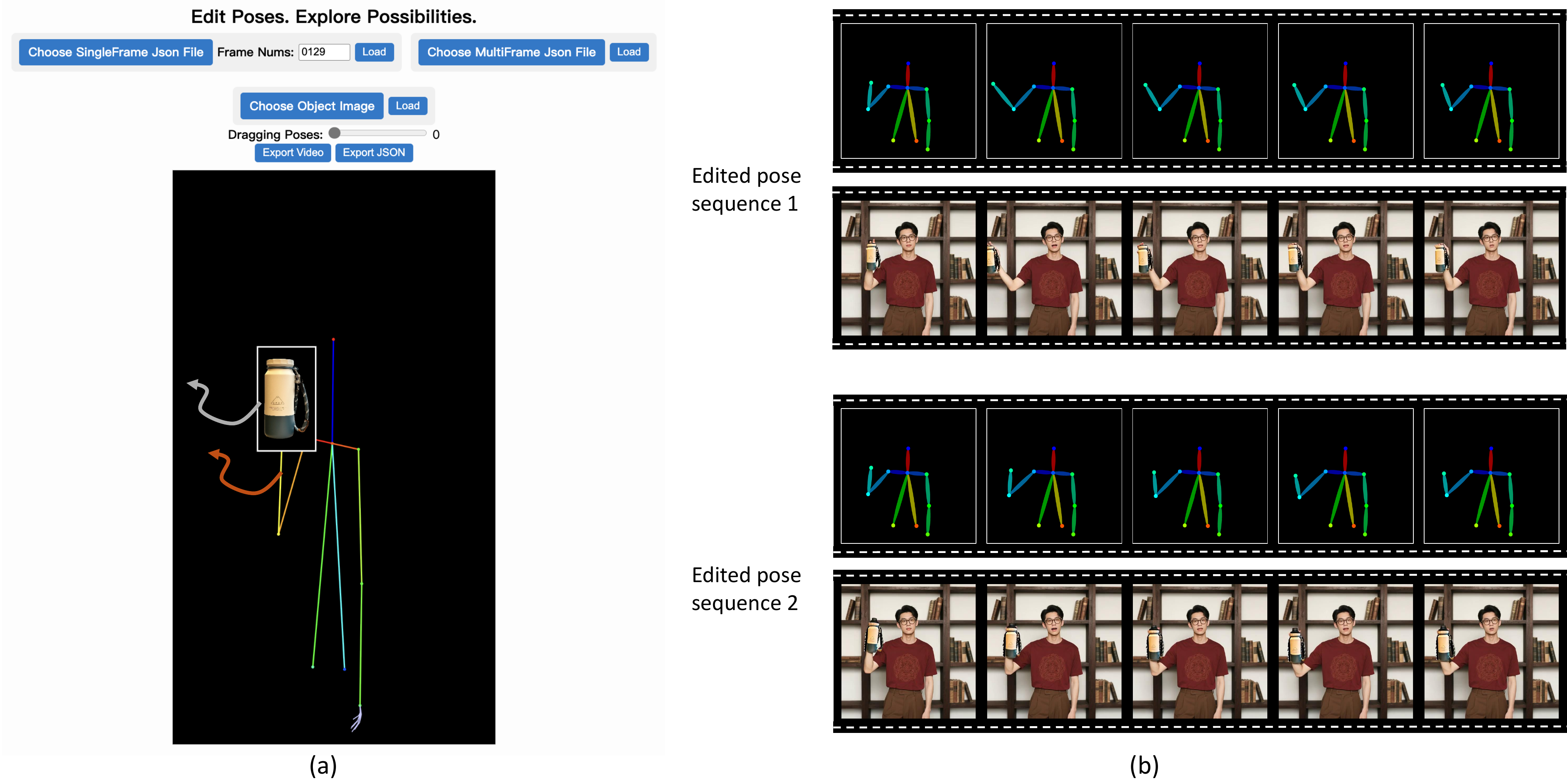}
    \caption{Interactive Demo. Our interactive demo supports uploading either a single frame or a sequence of human poses, along with an object image. It allows users to adjust the object size on the first frame by resizing the bounding box. Users can select specific frame indices to edit, and perform drag-and-drop editing of both the human pose and the object position on each selected frame. Additionally, the demo supports interpolation between edited frames to generate smooth motion sequences.}
    \label{fig:demo}
\end{figure*}

\subsection{Interactive Demo}
\label{sec:demo}
We provide an interactive web demo that allows users to intuitively define human-object interactions by manually dragging the human pose and object position. This interactive tool grants the model adaptive freedom to resolve ambiguities, bridging the gap between artistic intent and generative feasibility. The web UI and use examples are illustrated in Fig.~\ref{fig:demo}.

\vspace{-5px}
\section{Conclusion}
\label{sec:conclusion}
In this work, we propose HunyuanVideo-HOMA, a novel framework for human-object interaction (HOI) video generation under weak control conditions. 
We explore a weak HOI condition setting to alleviate the reliance on hard-to-obtain strong HOI motion inputs. To this end, we propose a context fusion strategy and the HOI Adapter leveraging the generative prior, enabling the generation of HOI videos with high generalization and natural human-object interactions under limited input conditions.
Extensive quantitative benchmarks and qualitative analyses demonstrate the superior performance of HunyuanVideo-HOMA compared to existing diffusion-based approaches in human-centric video generation and editing. Furthermore, we validate the generalizability and applicability of our framework across a variety of real-world scenarios, including various object appearance and motion, different types of poses, and a user-friendly web demo.

\clearpage
\bibliography{main}
\bibliographystyle{abbrvnat}

\appendix
\clearpage

\appendix
\section*{Appendix}

\section{Results of AnchorCrafter test set}
The quantitative results of the AnchorCrafter test set are shown in Table~\ref{tab:ac}. Similar to the results in the manuscript, ours achieves better results on FID, FVD, and OC, and get comparable results in other metrics. It should be noted that HOMA gets a higher score than AnchorCrafter, which requires target-object fine-tuning. This observation indicates the object preservation capability of our method.

\begin{figure*}[htbp]
    \centering
    \includegraphics[width=1.0\linewidth]{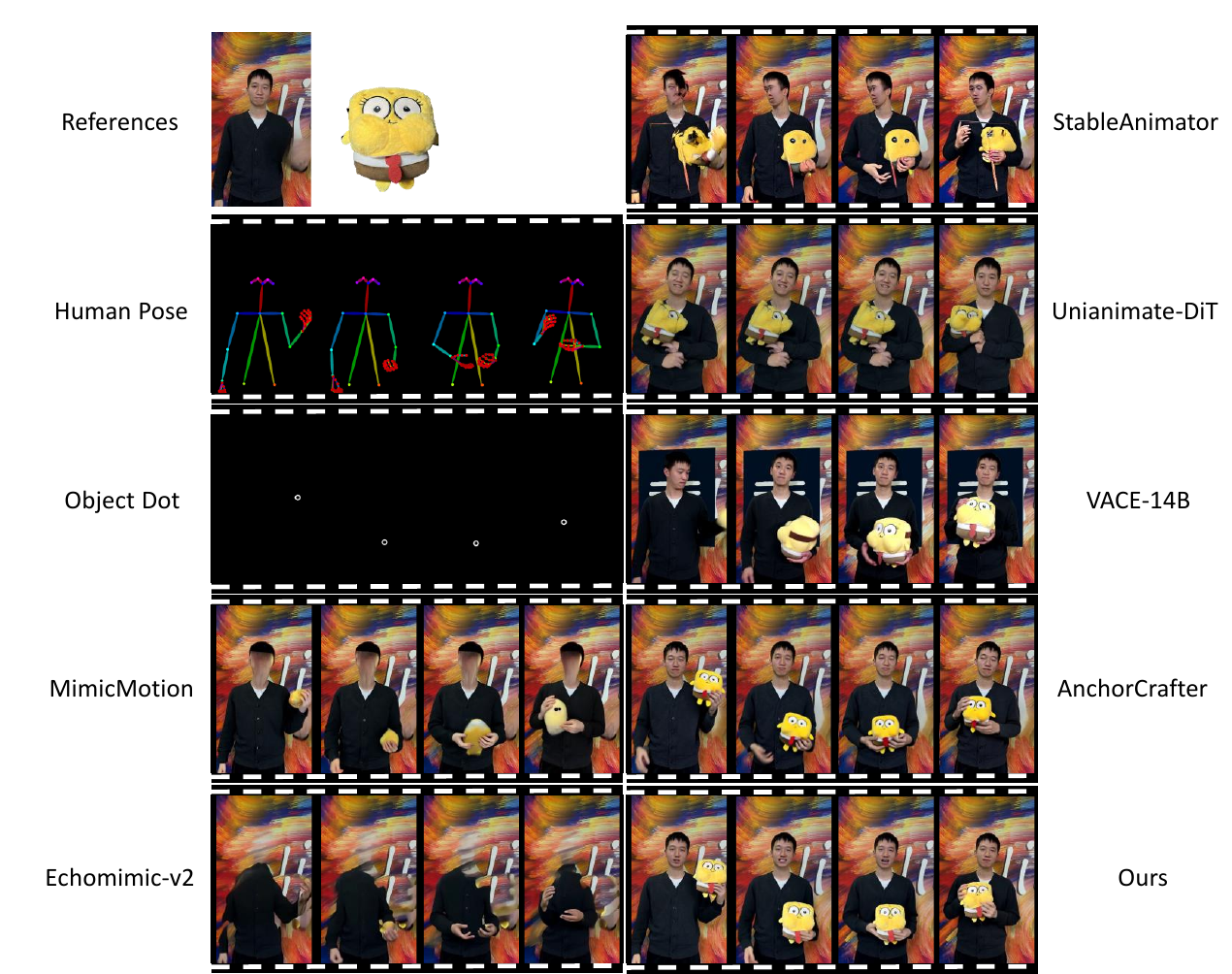}
    \caption{Qualitative results of AnchorCrafter test set.}
    \label{fig:expr_AC}
\end{figure*}

Qualitative results are illustrated in Fig.~\ref{fig:expr_AC}. 
Compared to other approaches, our method achieves better interactive motion while maintaining consistency for both the human and the object. In addition, unlike methods that require fine-tuning on the target object, our approach further enhances object consistency. We attribute this to the fact that AnchorCrafter relies on depth maps to control object motion. However, the estimated depth maps contain inaccuracies and tend to be overly smoothed, which can compromise the fidelity of the object's appearance. In contrast, our dot-trajectory-based control mechanism better preserves the object's visual characteristics without being affected by such artifacts.
\begin{table}[htbp]
  \centering
  \caption{Quantitative results of our method compared with SOTAs in AnchorCrafter test set. }
    \label{tab:ac}
  \resizebox{\linewidth}{!}{
  \begin{tabular}{@{}lccccccccc@{}}
    \toprule
  \multirow{2}{*}{Method} & \multicolumn{9}{c}{\textbf{AnchorCrafter test set}} \\
                                                              & FID↓            & FVD↓            & OC↑           &SC↑           &BC↑             & HAS↑                & HF↑                   & MS↑            & Sync-C↑ \\
    \midrule
    AnchorCrafter (w/ Finetuning)                             &  89.29          & 831.58          & 86.31         & 96.39        &  94.66         &  83.25              & 97.52                 &99.23            &  -\\
    MimicMotion                                               &  89.74          & 1262.72         & 83.86         & 93.48        &  93.36         & 88.60               &  99.25                &99.18            &  -\\
    Echomimic-v2                                              &  127.12         & 1840.43         & 79.33         &  93.77       & 92.45          &\textbf{88.61}       & 99.35                 &98.98            &  6.329\\
    StableAnimator                                            &  137.19         & 1649.72         & 80.87         & 90.02        & 90.86          & 83.52               & 98.31                 &98.16            &  -\\
    UniAnimate-DiT                                            &  85.12          & 859.61          & 84.03         &\textbf{97.29}& 95.23          & 82.67              & 96.74                  &\textbf{99.47}   & -\\
    VACE-14B                                                  &  85.10          & 1293.84         & 82.05         & 94.54        & 94.95          & 81.90               &\textbf{99.98}        & 98.82           &  -\\
    Ours                                                      &  \textbf{64.77} & \textbf{618.45} & \textbf{90.03}& 96.09        & \textbf{95.27} & 88.45               & 99.34                &99.24            &   4.864   \\
    \midrule
  \end{tabular}
  }
\end{table}

\section{Influence of the Copy Concat}
\begin{figure*}[htbp]
    \centering
    \includegraphics[width=\linewidth]{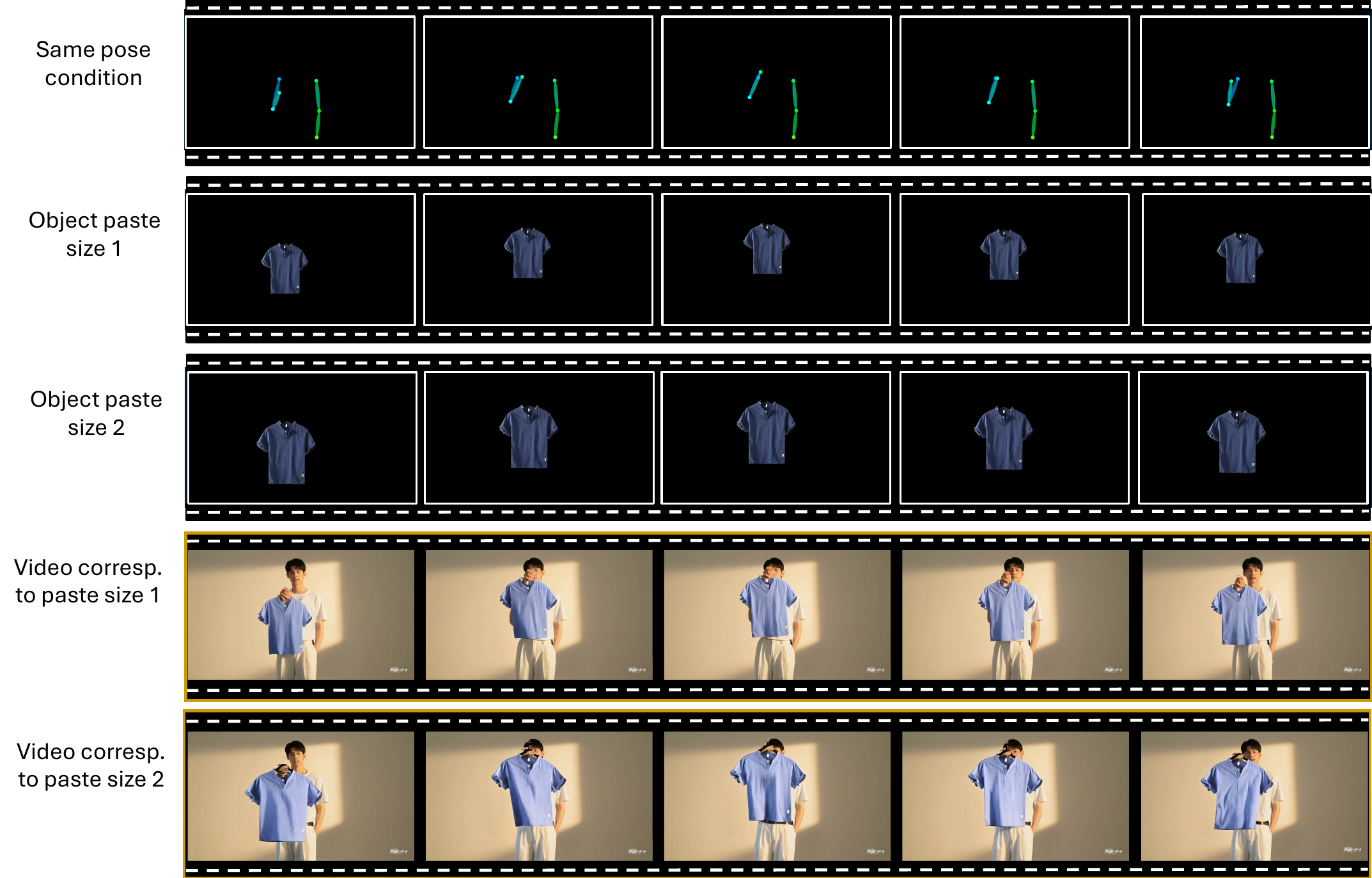}
    \caption{Influence of the copy concat.}
    \label{fig:copy-concat}
\end{figure*}

It is worth noting that although the dot trajectory itself does not constrain the object size, our pipeline introduces implicit control over object scale in two ways. First, the object image is channel-concatenated in the latent space, and the HOI Adapter module aligns the object region based on a mask, which enables a degree of size control. As illustrated in Fig.~\ref{fig:copy-concat}, different alignment sizes of the same object result in generated objects that follow the aligned scale. On the other hand, our interactive demo allows users to manually define the alignment size of the object, thereby providing additional flexibility for user-driven editing.

\section{Influence of MLLM feature}
\begin{figure}[htbp]
    \centering
    \includegraphics[width=\linewidth]{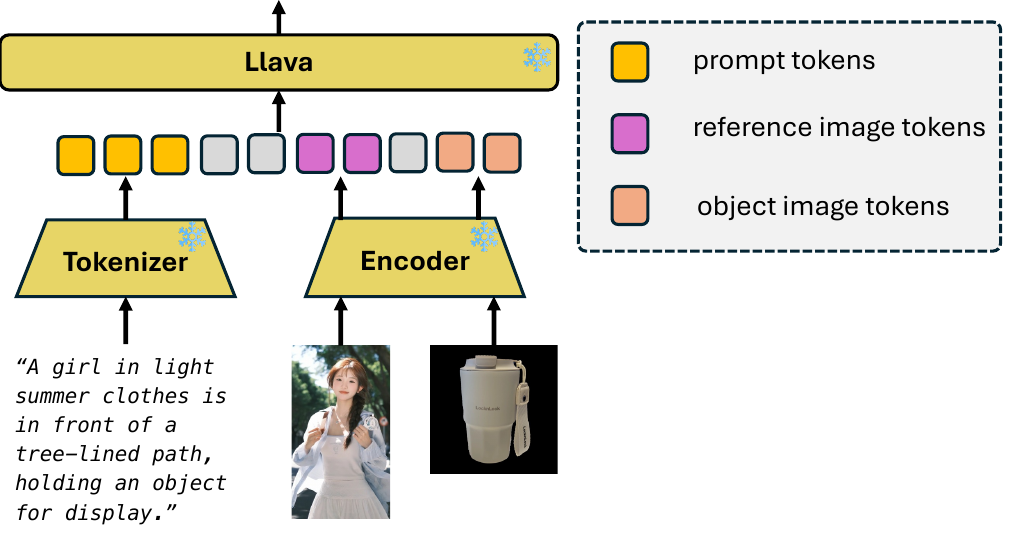}
    \caption{Network architecture of the MLLM.}
    \label{fig:mllm}
\end{figure}

The details of the MLLM we utilized are in Fig.~\ref{fig:mllm} It integrates both text prompts, human reference image, and object reference image as multimodal condition.
We analyze how these multimodal condition contribute to our model performance.

\begin{figure}[htbp]
    \centering
    \includegraphics[width=\linewidth]{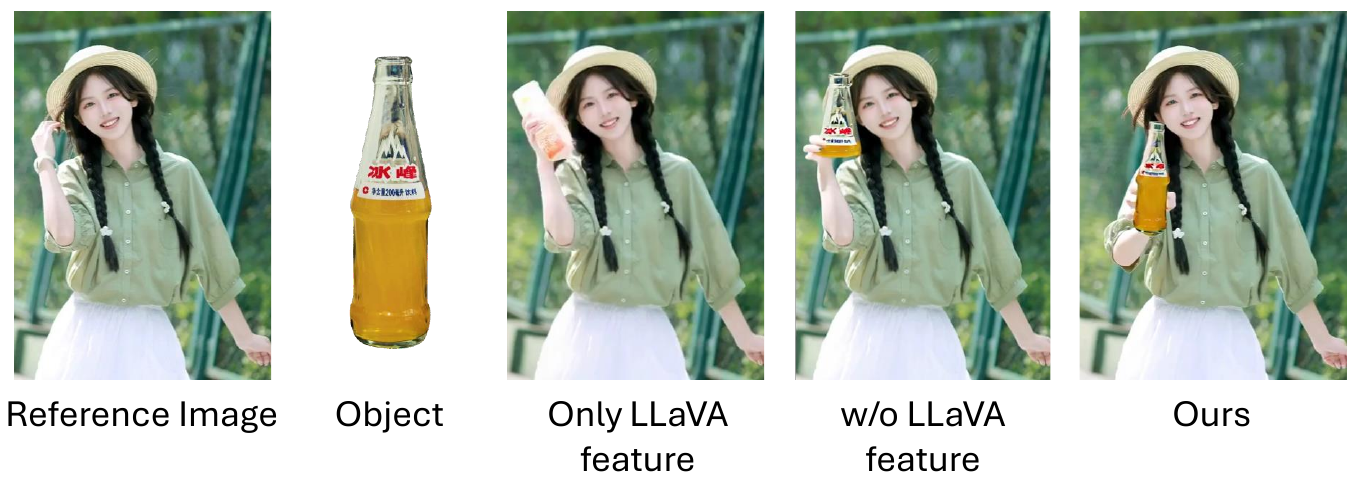}
    \caption{Influence of LLaVA feature.}
    \label{fig:llava}
\end{figure}

As shown in Fig.~\ref{fig:llava}, when using only text prompt and image tokens as input to guide object appearance, the model demonstrates a strong understanding of the object category and high-level semantics. However, it fails to accurately capture the low-level visual textures and detailed appearance, resulting in generated humans that conceptually recognize the object but exhibit incorrect or oversimplified surface details.

Conversely, when only adopting text prompt as condition, the model tends to misinterpret the object's semantic category, leading to structurally inconsistent or semantically irrelevant results. These findings highlight the limitation of relying solely on either semantic abstraction or context fusion for appearance injection.

This comparison underscores the necessity of combining both semantic and context fusion, as achieved by our proposed modules, which integrate high-level understanding with low-level fidelity to support more accurate and realistic HOI video generation.

\section{Variations of Different Trajectories}
\begin{figure*}[h]
    \centering
    \includegraphics[width=0.8\linewidth]{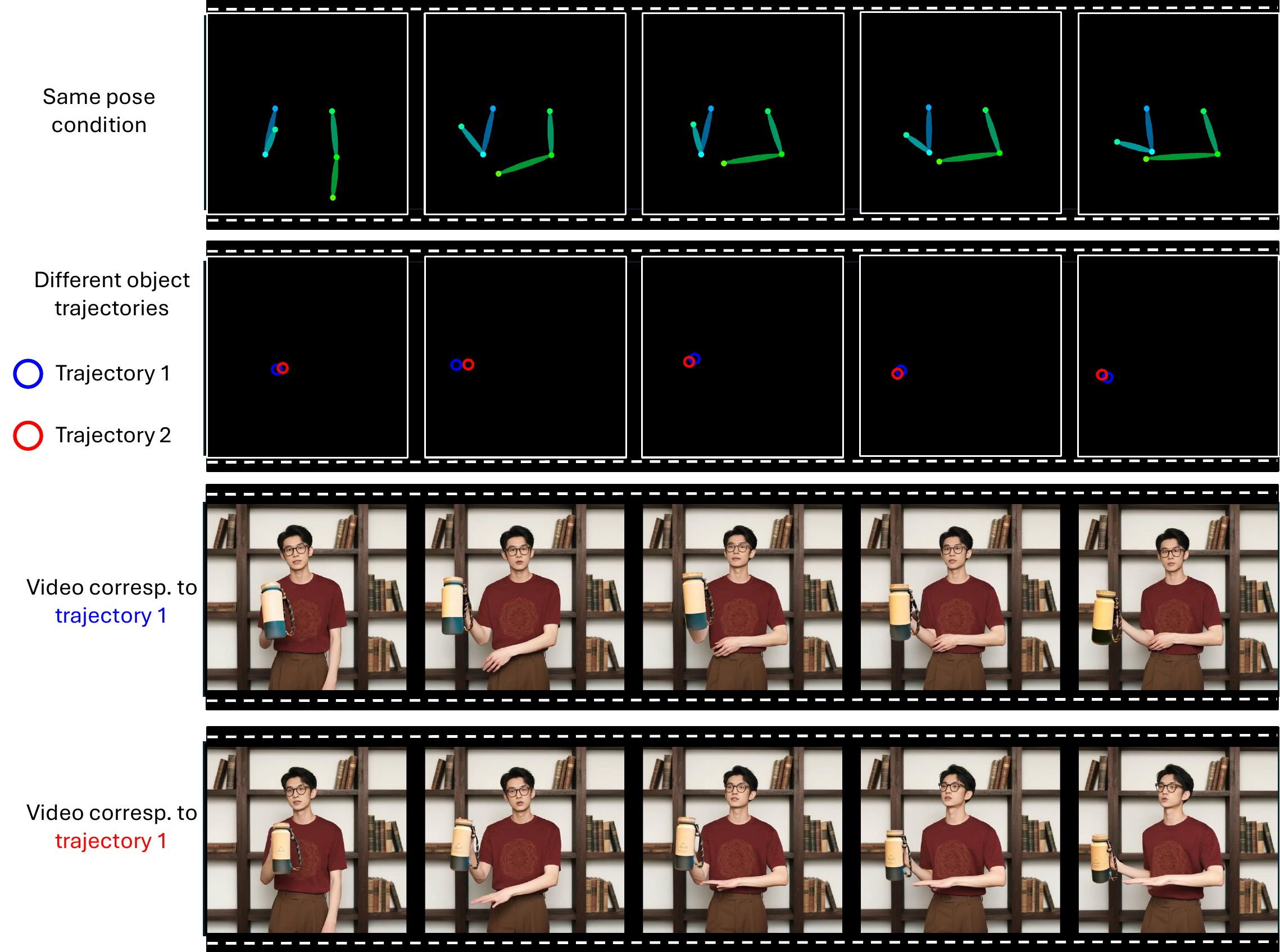}
    \caption{Variations of different trajectories.}
    \label{fig:diff-obj-traj}
\end{figure*}

We demonstrate the effect of varying object trajectories under the same human pose in Fig.~\ref{fig:diff-obj-traj}. By slightly modifying the motion of the object, our method enables diverse hand-object interaction patterns even with identical body poses. Notably, under weak control settings, the fine-grained hand motions adapt accordingly to the object’s trajectory, contributing to enhanced interaction diversity.

\section{Whole Data Pipeline}
\begin{figure}[htbp]
    \centering
    \includegraphics[width=\linewidth]{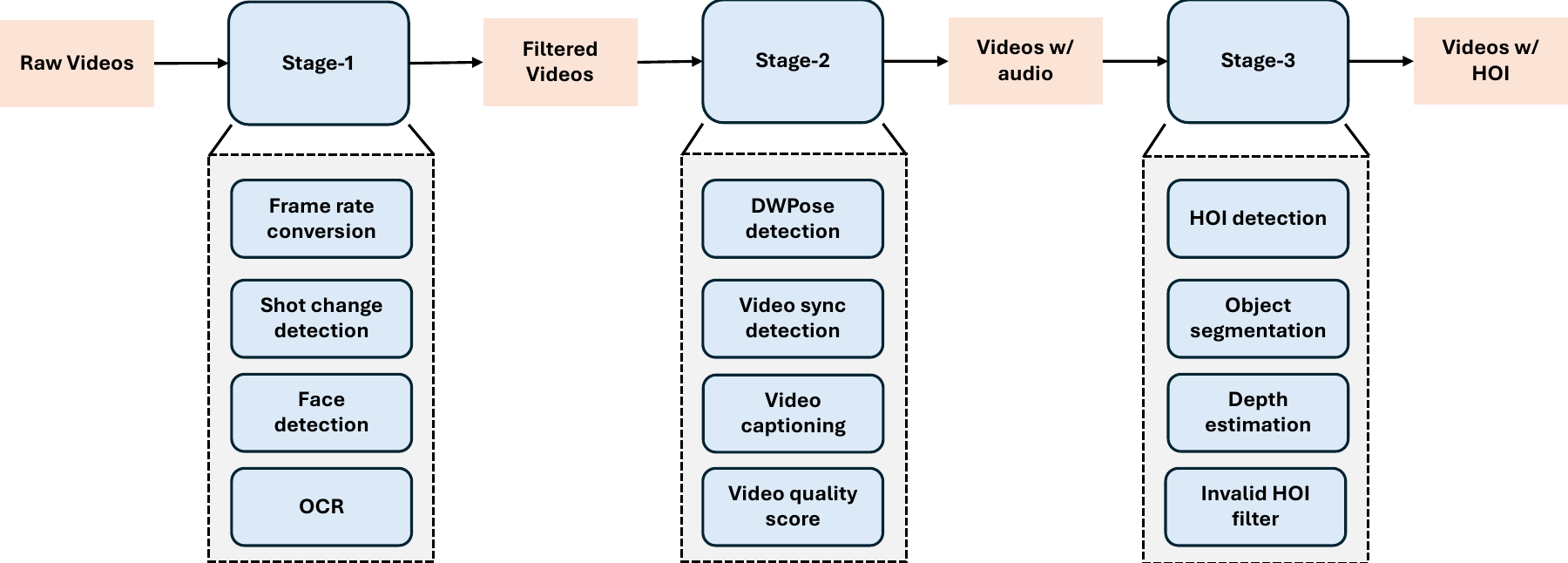}
    \caption{Whole data pipeline.}
    \label{fig:data-pipeline}
\end{figure}

To construct a high-quality dataset containing human-object interactions (HOI), we design a three-stage data processing pipeline that progressively filters, analyzes, and refines raw video content. The whole pipeline is shown in Fig~\ref{fig:data-pipeline}, where before the HOI filtering described in Sec.3.3, there are two stages to obtain a high-quality human video dataset. This modular pipeline ensures that the resulting data is both semantically meaningful and structurally consistent, providing a reliable foundation for downstream HOI-related tasks.

\textbf{Stage 1: Preprocessing and Basic Filtering.}
Raw videos are first passed through a series of preprocessing steps to remove low-quality or irrelevant content. These include frame rate conversion to standardize temporal resolution, shot change detection to segment video scenes, face detection to ensure the presence of humans, and optical character recognition (OCR) to filter out videos with overlaid text (e.g., subtitles or watermarks). The output is a collection of filtered videos suitable for further analysis.

\textbf{Stage 2: Semantic Analysis and Quality Assessment.}
The filtered videos are then enriched with semantic and quality-related metadata. We employ DWPose to extract 2D human pose keypoints, followed by a synchronization check between video and audio streams to ensure temporal alignment. Video captioning is applied to generate textual descriptions for content understanding, and a visual quality scoring model is used to assign a quality score to each video. This stage outputs videos with verified human presence, audio consistency, and sufficient visual-semantic quality.

\textbf{Stage 3: HOI Detection and Refinement.}
In the final stage, we focus on identifying valid HOI instances. A dedicated HOI detection model is applied to locate interaction events between humans and objects. Subsequently, object segmentation is used to accurately localize the interacted object, and depth estimation provides geometric context for interaction understanding. To ensure dataset integrity, an additional invalid HOI filtering step removes clips with ambiguous or low-confidence interactions. The resulting dataset comprises videos with rich, valid, and clearly defined HOI content.

This pipeline enables robust, scalable video data curation and supports reliable training and evaluation of models for HOI understanding and generation.

\section{Long Video Generation}
To accomplish long video generation, we utilize a training-free method operated on the latent space similar to MimicMotion. The input is divided into overlapping segments. Each segment is denoised independently using the same references and corresponding control signal segments. To maintain temporal consistency across segments, overlapping frames are smoothly fused based on their position, with closer frames receiving higher weights. After all denoising steps, segments are merged to produce a coherent long video sequence. We highlight that our context fusion and adapter design can be directly applied to this long-video inference strategy, enabling strong consistency in both motion and appearance across segments. Long video results are in the video demos.

\section{Limitations}
\label{sec:limitation}
\begin{figure}[tbp]
    \centering
    \includegraphics[width=\linewidth]{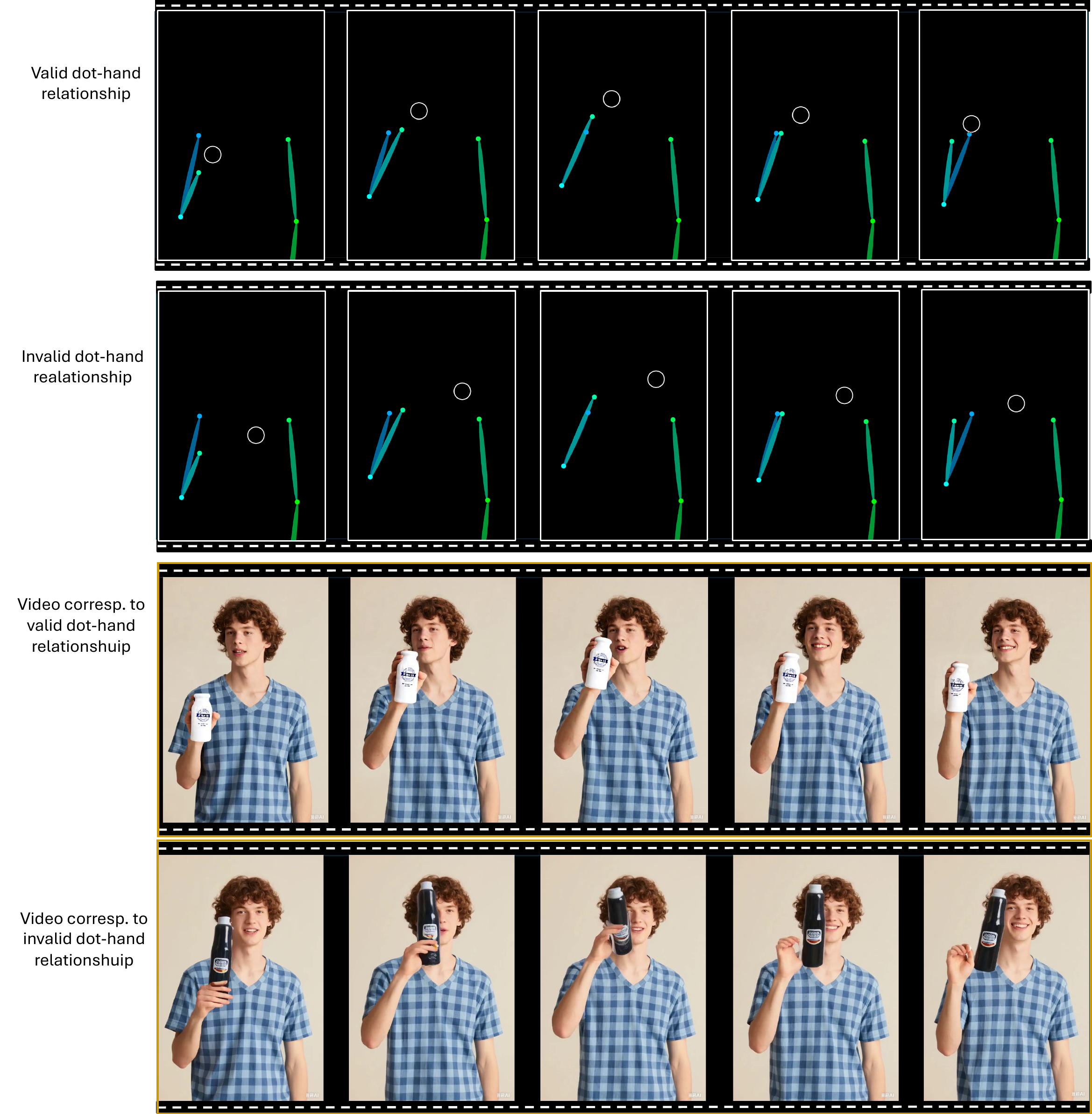}
    \caption{Limitation.}
    \label{fig:limit}
\end{figure}

We observe that the distance between the dot and the hand significantly influences the final generation quality. As shown in Fig.~\ref{fig:limit}, when the dot is far from the hand, the generated object tends to exhibit reduced consistency and interaction plausibility. Nevertheless, even in such challenging cases, the model still strives to produce the most reasonable human-object interaction as it can, reflecting its strong generative capability.
Moreover, we highlight that, due to the generalizability of our weak condition framework, such deviations can be effectively corrected via the interactive demo as discussed in Sec.4.6 which allows users to refine and reposition the input points accordingly.

\clearpage

\end{document}